%% file: Kurz-SphericalHarmonics.tex
\documentclass[preprint,1p,11pt]{StyleFiles/ISAS_IR}

\pdfoutput=1

\usepackage{amsmath}
\usepackage{amssymb}
\usepackage[utf8]{inputenc}
\usepackage[keeplastbox]{flushend}
\usepackage{graphicx}
\usepackage{siunitx}
\usepackage{subcaption}

\usepackage{gkmacros}

\usepackage{mathdots} %for iddots

\input{StyleFiles/Abbreviations}

\input{StyleFiles/Defs}
\input{StyleFiles/Format}

\title{\LARGE \bf
Three-dimensional Simultaneous\\ Shape and Pose Estimation for Extended Objects\\ Using Spherical Harmonics
}

\author{Gerhard Kurz, Florian Faion, Florian Pfaff, Antonio Zea, and Uwe D. Hanebeck% <-this % stops a space
\address{Intelligent Sensor-Actuator-Systems Laboratory (ISAS)\\
	Institute for Anthropomatics and Robotics,\\
Karlsruhe Institute of Technology (KIT), Germany \\
e-mail: {\tt kurz.gerhard@gmail.com, florian.faion@gmail.com, florian.pfaff@kit.edu, antonio.zea@kit.edu, uwe.hanebeck@ieee.org}.}% 
}

\begin{document}
	
\begin{frontmatter}

\begin{abstract}
	We propose a new recursive method for simultaneous estimation of both the pose and the shape of a three-dimensional extended object. The key idea of the presented method is to represent the shape of the object using spherical harmonics, similar to the way Fourier series can be used in the two-dimensional case. This allows us to derive a measurement equation that can be used within the framework of nonlinear filters such as the UKF. We provide both simulative and experimental evaluations of the novel techniques.
\end{abstract}
\end{frontmatter}

\section{Introduction}
We consider the problem of simultaneously estimating the position and shape of an extended object. This problem is very widespread in robotics and autonomous navigation whenever a robot encounters unknown objects whose shape is not known a priori. 

The problem of extended object tracking has been considered in literature by a number of authors~\cite{Fusion15_Faion,JAIF15_Faion,Fusion11_Baum-RHM,IROS11_Baum,wahlstrom2015,feldmann2011,hirscher2016}.
In some cases, the shape of the object is assumed to be known a priori, in other cases certain parameters (e.g., the radius of a circle, the length of the major and minor axis of an ellipse) are assumed to be unknown, and sometimes the entire shape of the object has to be estimated. One of the key challenges in extended object tracking is the question of how to associate measurements stemming from the object to a particular measurement source on the object. Common solutions are the greedy association model (GAM)~\cite{Fusion14_Faion} that associates each measurement to the most likely source and the spatial distribution model (SDM)~\cite{gilholm2005} that assumes a probability distribution of how likely each possible source is. 

It is not only desirable to track the movement of the object but to estimate its shape as well, thus creating a model of the observed object. The techniques presented in this paper can even be applied to nonrigid objects whose shape is subject to changes over time. Another application of the proposed methods consists in the classification of objects, where objects in the environment can be identified based on their estimated shape by comparing with a database of known objects. Moreover, the shape of an object can be used for grasp planning in robotic manipulation.
%Furthermore, it would even be possible to perform unsupervised learning and cluster different objects depending on their shape.

The main deficiency of most approaches in literature is that they are limited to two-dimensional problems. To address this limitation, we propose a three-dimensional method in this paper.

\begin{figure}
	\centering
	\begin{subfigure}{.46\linewidth}
		\centering
		\includegraphics[height=50mm]{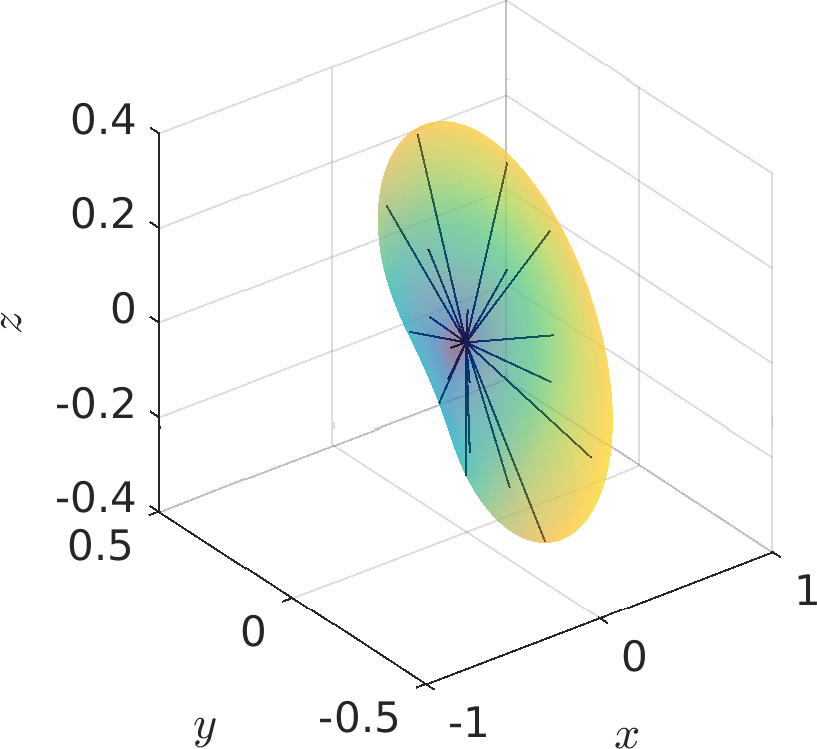}
		\caption{3D Shape in Cartesian coordinates.}
	\end{subfigure}
	\begin{subfigure}{.46\linewidth}	
		\centering
		\includegraphics[height=50mm]{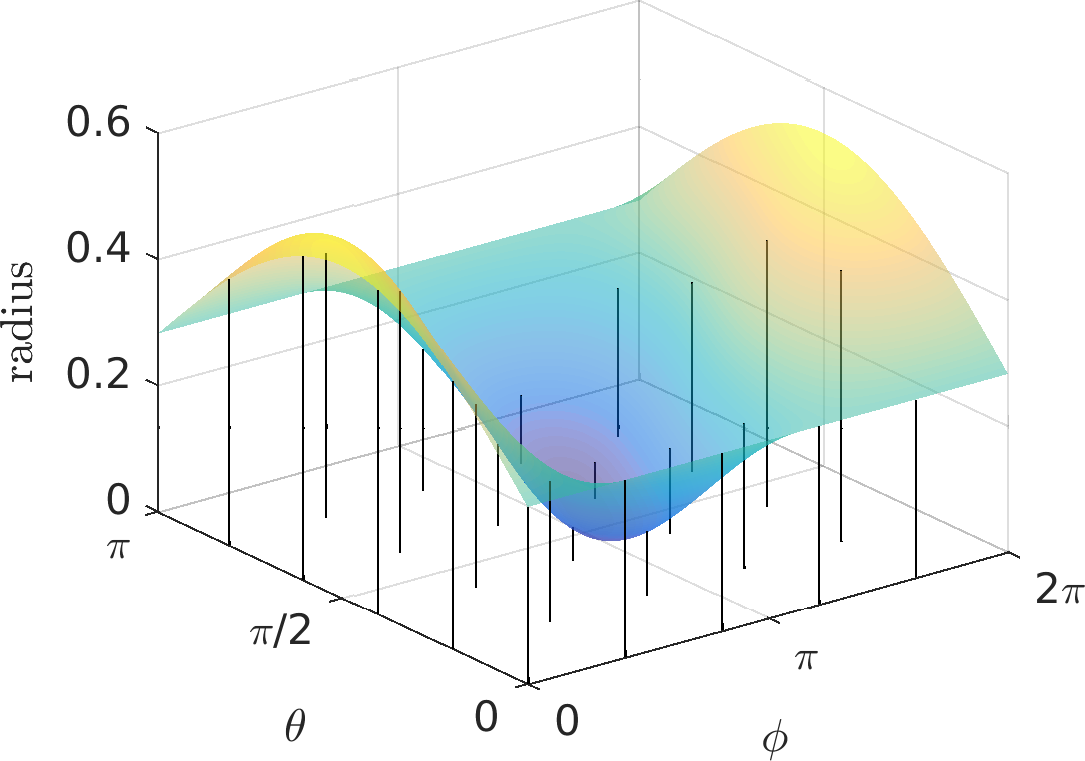}	
		\caption{Extent in spherical coordinates.}
	\end{subfigure}	
	\caption{Star-convex shape in 3D represented by spherical harmonics. The function on the right maps spherical coordinates to the extent of the object in the corresponding direction.} 
	\label{fig:starconvex}
\end{figure}

%todo difference to 3D scanning: recursive, few measurements per time step, high noise, no manual post processing

When estimating the shape of an extended object, a key question is the choice of the representation of the shape. In literature, Fourier series \cite{Fusion11_Baum-RHM}, different types of splines~\cite{JAIF14_Kurz,Fusion16_Zea}, as well as Gaussian Processes \cite{dragiev2011,wahlstrom2015,hirscher2016,oezkan2016} have been proposed for the 2D case. In the following, we will rely on a three-dimensional generalization of the Fourier series approach. This approach---in 2D as well as in 3D---represents star-convex\footnote{A set is star-convex if there exists a star point inside the set such that for every point in the set the line connecting this point to the star point is entirely contained within the set.} shapes by defining a function that maps every direction to the distance from the so-called star point to the boundary of the object along that direction. An illustration of this concept can be seen in Fig.~\ref{fig:starconvex}. Another 3D approach in literature is the extrusion random hypersurface model proposed in \cite{SDF14_Zea}. However, this method is limited to estimation of objects that can be well represented using (scaled) extrusions, such as rotationally symmetric objects. Furthermore, there is an approach based on extended Gaussian images by Sun et al.~\cite{sun2014} that can in principle be generalized to the three-dimensional case.

The key idea in this paper consists in representing the shape of an extended object using so-called spherical harmonics. Spherical harmonics can be seen as a counterpart of Fourier series defined on the unit sphere \cite{atkinson2012,mueller1966,groemer1996,macrobert1948}, \cite[Sec.~4]{chisholm1976}. Thus, they allow approximation of arbitrary functions defined on the sphere and can be used to elegantly represent the shape of the object under consideration. As a result, estimating the shape is equivalent to estimating the parameters of the weights of the different basis functions. Spherical harmonics have previously been used  as a representation of spherical probability distributions~\cite{MFI17_Pfaff-SphericalHarmonics}, as rotation invariant shape descriptors~\cite{kazhdan2003}, and for view-based robot localization~\cite{friedrich2007}.
%\cite[Appendix~A]{kreutzmann2013},

\section{Spherical Harmonics}
Spherical harmonics are special functions on the surface of a sphere and resemble Fourier series on the circle. In this section, we will give a concise introduction into the topic of spherical harmonics. %Note that there are a number of different conventions in literature, which differ in normalization constants or sign.

\begin{figure*}
	\centering
	\newcommand{\bwidth}{17mm}
	\newcommand{\spacer}{\hspace{5mm}}
	\setlength{\lineskip}{1mm}
	\setlength{\tabcolsep}{1mm}	
	\begin{tabular}{cccccccccc}
		& $m=-4$ & $m = -3$ & $m = -2 $ & $ m=-1$ & $m=0$ & $m=1$ & $m=2$ & $m=3$ & $m=4$ \\
		\rotatebox{90}{\spacer$l=0$} & & & & & \includegraphics[width=\bwidth]{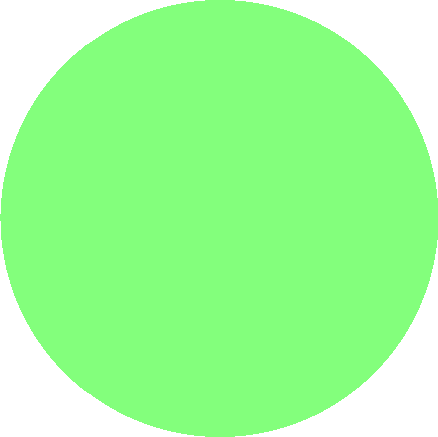} & & & & \\
		\rotatebox{90}{\spacer$l=1$} & & & & \includegraphics[width=\bwidth]{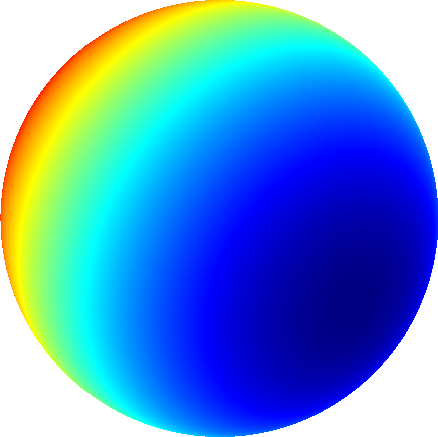} &	
		\includegraphics[width=\bwidth]{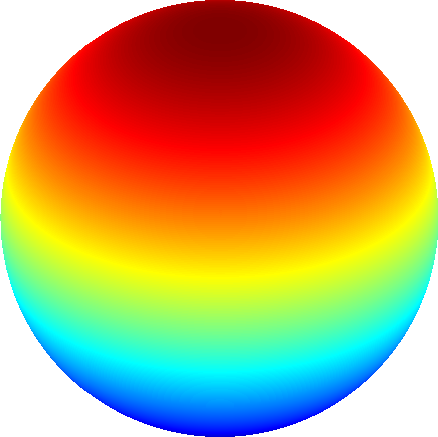} &	
		\includegraphics[width=\bwidth]{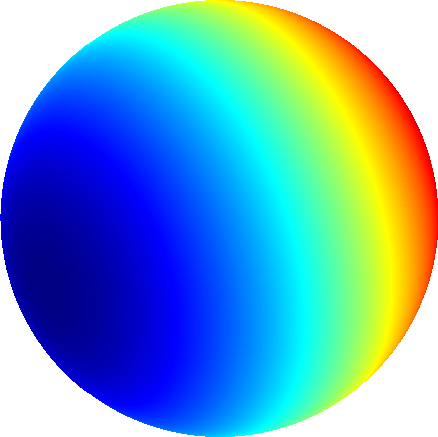} & & &\\
		\rotatebox{90}{\spacer$l=2$} & & &
		\includegraphics[width=\bwidth]{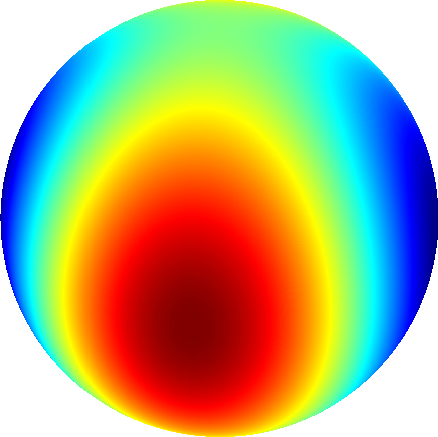} &		
		\includegraphics[width=\bwidth]{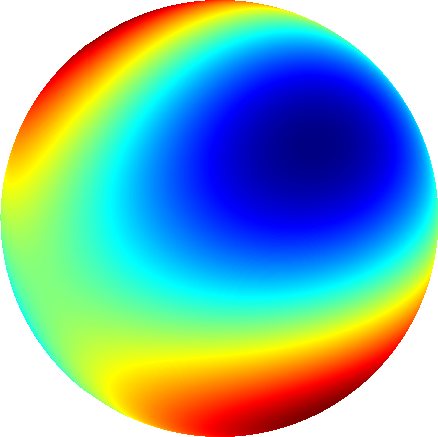} &	
		\includegraphics[width=\bwidth]{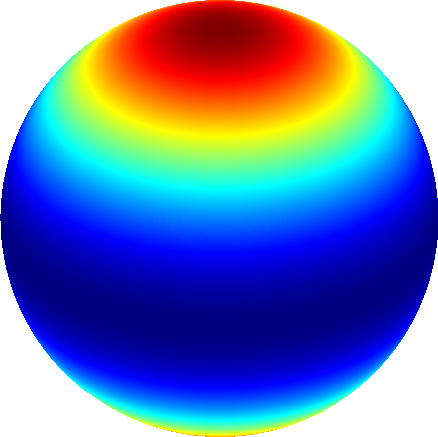} &	
		\includegraphics[width=\bwidth]{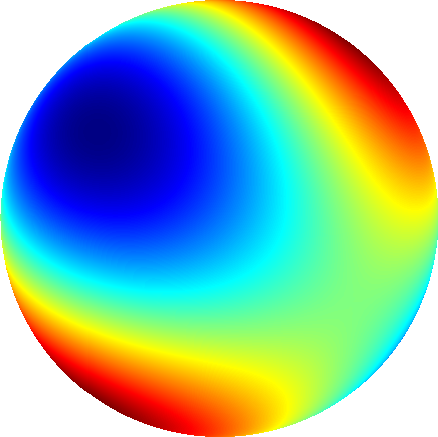} &
		\includegraphics[width=\bwidth]{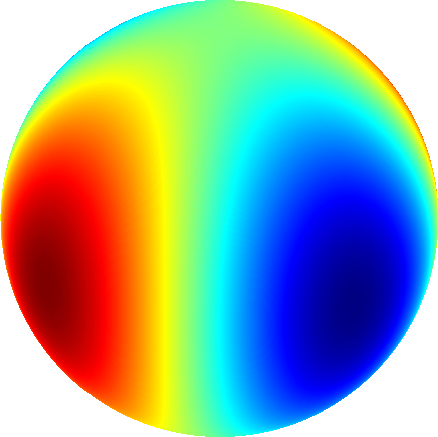} & & \\	
		\rotatebox{90}{\spacer$l=3$} & & \includegraphics[width=\bwidth]{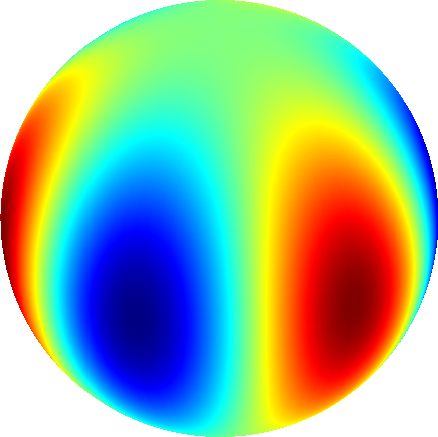} &		
		\includegraphics[width=\bwidth]{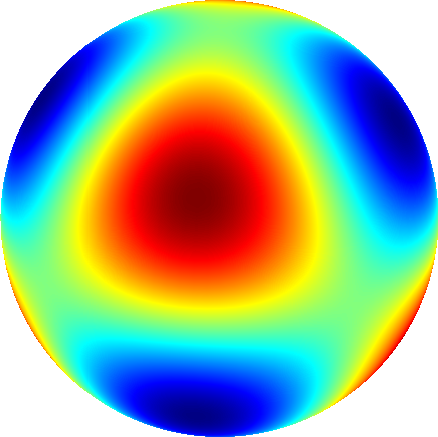} &	
		\includegraphics[width=\bwidth]{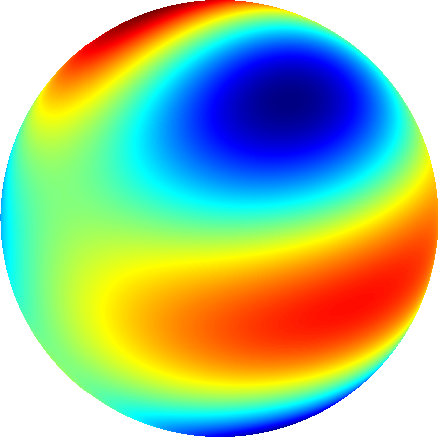} &	
		\includegraphics[width=\bwidth]{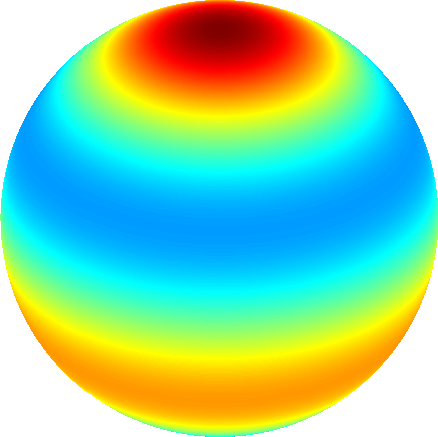} &	
		\includegraphics[width=\bwidth]{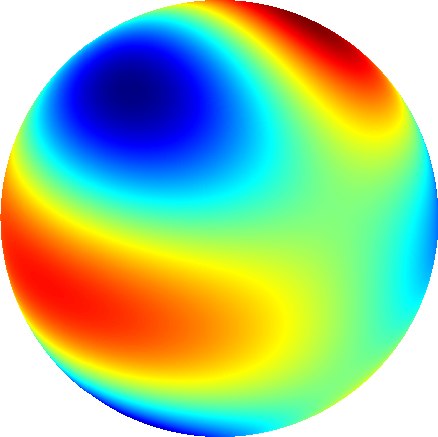} &
		\includegraphics[width=\bwidth]{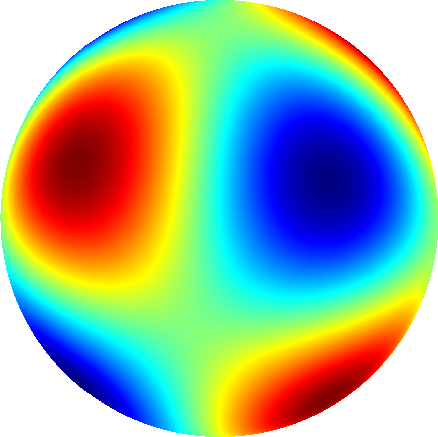} &
		\includegraphics[width=\bwidth]{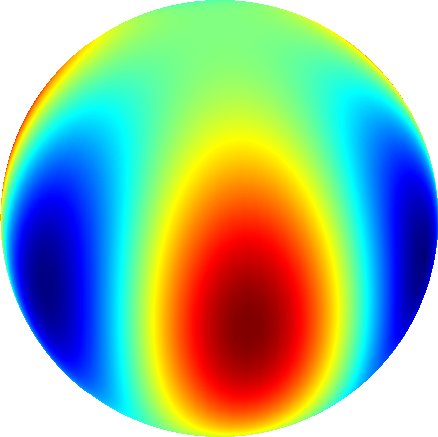} & \\
		\rotatebox{90}{\spacer$l=4$} & \includegraphics[width=\bwidth]{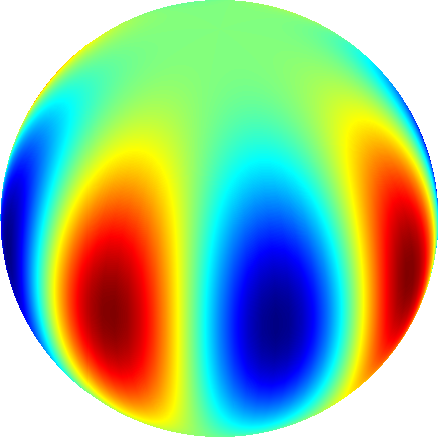} &
		\includegraphics[width=\bwidth]{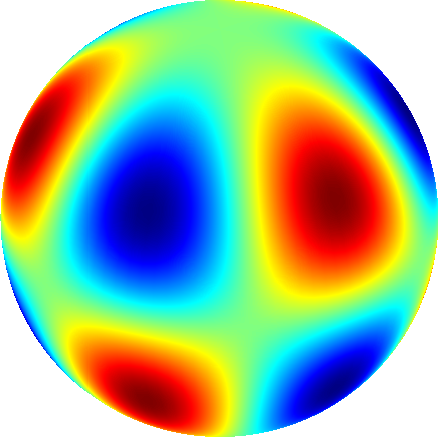} &
		\includegraphics[width=\bwidth]{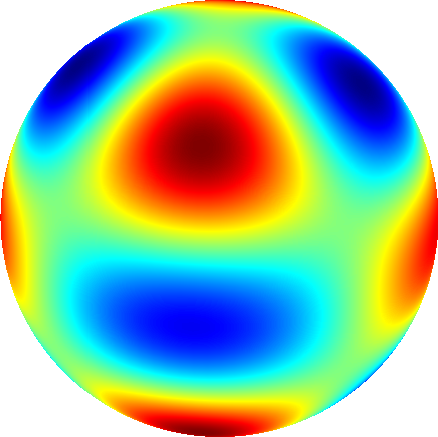} &	
		\includegraphics[width=\bwidth]{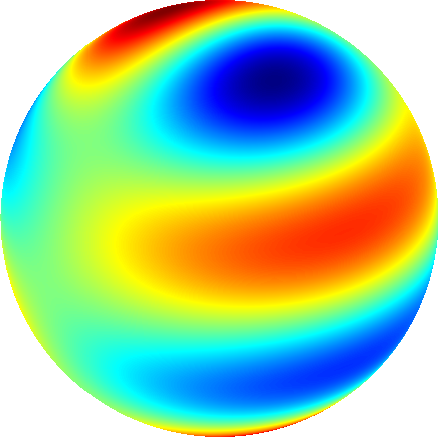} &
		\includegraphics[width=\bwidth]{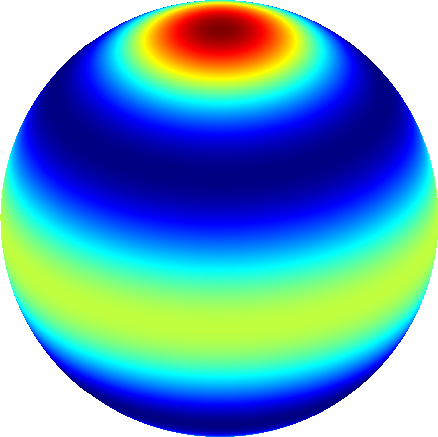} &
		\includegraphics[width=\bwidth]{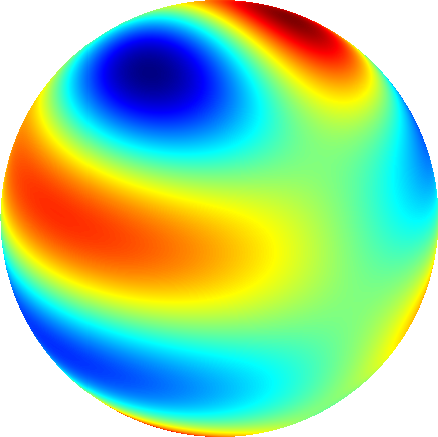} &
		\includegraphics[width=\bwidth]{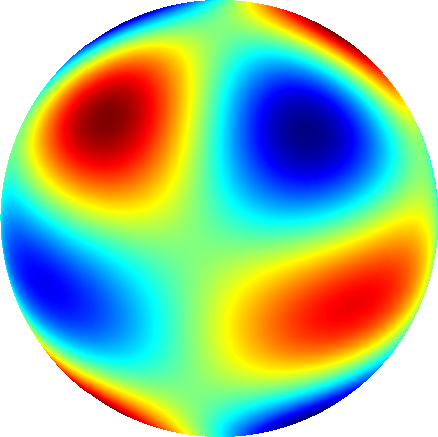} &
		\includegraphics[width=\bwidth]{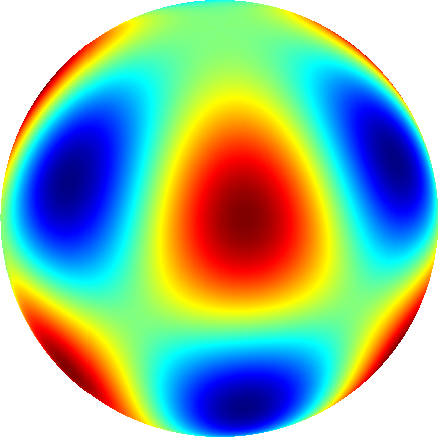} &
		\includegraphics[width=\bwidth]{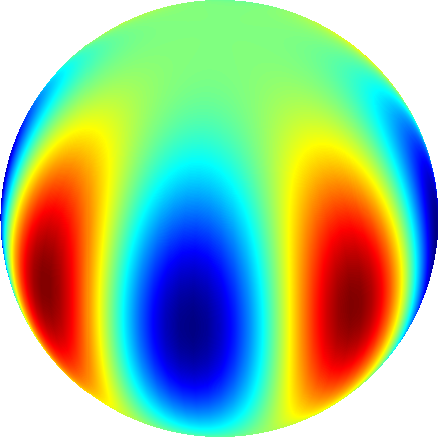} \\	
	\end{tabular}
	\caption{Basis functions for spherical harmonics. Positive values are depicted in red, negative values in blue.}	
	\label{fig:basisfunctions}
\end{figure*}

\subsection{Basis Functions}
First, we introduce the basis functions that are defined on the sphere. In this paper, we use spherical coordinates according to the following convention
\begin{align*} %MATLAB Convention, see https://de.mathworks.com/help/matlab/ref/sph2cart.html
	x &= r \sin(\theta) \cos(\phi) ,\\ 
	y &= r \sin(\theta) \sin(\phi) , \\
	z &= r \cos(\theta) ,
\end{align*}
where $\phi \in [0,2\pi), \theta \in [0,\pi], r\geq0$.

There are various different conventions for the definition of spherical harmonics, which mostly differ by constant factors and sign. We use the definition by Blanco et al. \cite{blanco1997}, which in turn is based on Chisholm's work \cite{chisholm1976}.

There are complex and real versions of spherical harmonics. The basis functions for complex spherical harmonics are defined as \cite[eq.~(1)]{blanco1997}
\begin{align*}
	Y_l^m (\theta, \phi) = (-1)^m N_l^m P_l^m(\cos(\theta)) \exp(i m \phi)
\end{align*}
with constants $ N_l^m $ and associated Legendre polynomials $P_l^m(\cdot)$, where $l \in \mathbb{N}_0,  m \in \mathbb{Z}, -l \leq m\leq l$. This yields a triangular structure as illustrated in Fig.~\ref{fig:basisfunctions}.
%\begin{align*}
%	\begin{array}{ccccccc}
%		& & & Y_0^0 \\
%		& & Y_1^{-1} & Y_1^{0} & Y_1^{1} \\
%		& Y_2^{-2} & Y_2^{-1} & Y_2^{0} & Y_2^{1} & Y_2^{2} \\
%		\iddots & \vdots & \vdots & \vdots & \vdots & \vdots & \ddots \\
%	\end{array}
%\end{align*}
Thus, the number of basis functions up to order $L$ is $(L+1)^2$.
The constant factors $N_l^m$ are given according to \cite[eq.~(2)]{blanco1997}
\begin{align*}
	N_{l}^m = \sqrt{\frac{2l+1}{4\pi} \cdot \frac{(l-m)!}{(l+m)!}} \ .
\end{align*}
Furthermore, the associated Legendre polynomials $P_l^m$ \cite[Ch.~8]{abramowitz1972} are defined according to \cite[eq.~(3)]{blanco1997}
\begin{align*}
	P_l^m(x) = \frac{1}{2^l \, l!} (1-x^2)^{m/2} \frac{\partial^{l+m}}{(\partial x)^{l+m}} (x^2-1)^l \ .
\end{align*}

Following \cite[eq.~(6)]{blanco1997}, we now define the real spherical harmonics according to
\begin{align*}
	S_l^m(\theta, \phi) = \begin{cases}
	\frac{(-1)^m}{\sqrt{2}} (Y_l^m (\theta, \phi) + \overline{Y_l^m(\theta, \phi)}),  & m > 0 \\
	Y_l^0 (\theta, \phi), & m = 0 \\
	\frac{(-1)^m}{i \sqrt{2}} (Y_l^{-m} (\theta, \phi) - \overline{Y_l^{-m}(\theta, \phi)}), & m < 0
	\end{cases} \ ,
\end{align*}
where $\overline{a}$ refers to the complex conjugate of a complex number $a \in \mathbb{C}$. This expression can be simplified according to 
\begin{align*}
	S_l^m(\theta, \phi) = \begin{cases}
		N_l^m P_l^m(\cos(\theta)) \sqrt{2} \cos(m\phi),  & m > 0 \\
		N_l^0 P_l^0(\cos(\theta)),  & m = 0 \\		
		N_l^{-m} P_l^{-m}(\cos(\theta))\sqrt{2} \sin(-m\phi),  & m < 0 \\				
	\end{cases} \ .
\end{align*}
These basis functions are illustrated in Fig.~\ref{fig:basisfunctions} for $l=0,\dots, 4$.

%definition of Legendre polynomials $P_l^m$ \cite[Ch.~8]{abramowitz1972}
%
%according to \cite[Sec.~A.1.1]{kreutzmann2013}
%%see also https://de.mathworks.com/help/matlab/ref/legendre.html#f89-1002472
%\begin{align*}
%	P_l(x) = \frac{1}{2^l l!} \frac{d^l}{(dx)^l} (x^2-1)^l
%\end{align*}
%
%associated Legendre polynomials
%\begin{align*}
%	P_l^m(x) = (-1)^m (1-x^2)^{m/2} \frac{d^m}{(dx)^m} P_l(x)
%\end{align*}
%definition of this is wroing in kreutzmann2013 because m-th derivative is missing
%$(-1)^m$ is sometimes omitted (Condon--Shortley phase)

\subsection{Series Representation}
Using the basis functions defined above, we can use a series representation to approximate functions on the unit sphere. This is similar to the way a Fourier series approximates periodic functions using cosine and sine basis functions of different frequencies. 

The real-valued series representation using spherical harmonics basis functions is given by
\begin{align*}
	f(\theta, \phi) = \sum_{l=0}^{L} \sum_{m=-l}^l w_l^m  S_l^m(\theta, \phi) \ ,
\end{align*} 
i.e., each basis function is weighted with a coefficient $w_l^m$. Thus, a star-convex shape in $\mathbb{R}^3$ with star point $[0,0,0]^T$ can be represented in spherical coordinates with $f(\theta, \phi)$.
%\footnote{A set $S\subseteq \mathbb{R}^n$ is star convex if there exists a star point $s \in S$ such that for every $s' \in S$ the line segment from $s$ to $s'$ is contained in $S$.}

In order to achieve an efficient implementation, it is possible to drop all constants from $S_l^m(\theta, \phi)$, i.e., all terms that only depend on $l$ and $m$ and include them in the weight $w_l^m$. In that case, the uncertainty of the weighting parameter needs to be scaled accordingly, i.e., its covariance matrix must be divided by the square of the omitted constants.

\section{Bayesian Extended Object Tracking}
The system state $\vecx_k$ at time step $k$ contains the parameters determining the shape of the extended object under consideration as well as further information, such as its position. The shape is represented using the weighting coefficients of the spherical harmonics series. Thus, we can define the state of an object at position $[p_1, p_2, p_3]^T$ as
$$ \vecx = [p_1, p_2, p_3, w_0^0, w_1^{-1}, w_1^0, w_1^1, \dots  ]^T \ , $$
where we omit the time index $k$. Depending on the application, it is possible to augment the state with other quantities such as orientation, velocity, or angular velocity.

\subsection{System Model}
The state vector evolves according to a system model
$$ \vecx_{k+1} = \veca_k(\vecx_k,\vecu_k) + \vecw_k \ , $$
where $\vecw_k$ is zero-mean Gaussian noise. The most basic model is a random walk model where $\veca_k(\vecx_k, \vecu_k) = \vecx_k$. If the object is rotating at a known angular velocity, this can be modeled as an input $\vecu_k$, which leads to a nonlinear system model.
Here, the rotation function $\veca_k(\cdot, \cdot)$ performs the rotation by computing new spherical harmonics coefficients as described in \cite{blanco1997}. This method is similar to shifting a Fourier series by changing its phase.

If the angular velocity is not known, it is also possible to include it into the state vector and to estimate it along with the other parameters. Note, however, that this entails a certain ambiguity because a change in orientation can be misconstrued as a change in shape and vice versa.

\subsection{Measurement Model}
The measurement equation can be derived using the spherical harmonics series representation. First, we shift the measurement $\vecy_k$ by $[p_1, p_2, p_3]^T$ to obtain $\vecz_k = \vecy_k - [p_1, p_2, p_3]$. Then, we convert the coordinates of $\vecz_k$ into polar coordinates. Finally, we obtain the equation
$$  \vecz = \underbrace{\frac{\vecz}{\Vert \vecz \Vert} \sum_{l=0}^{L} \sum_{m=-l}^l w_l^m  S_l^m(\theta_z, \phi_z)}_\text{measurement source} + \, \vecv \ , $$
where we omit the time indices for improved readability. Here, $\theta_z$ and $\phi_z$ are polar coordinates of $\vecz$ and $\vecv$ is zero-mean Gaussian noise. The labeled term corresponds to the measurement source, i.e., the point on the objects surface that the measurement is assumed to originate from based on the greedy association model. Observe that $\vecz$ occurs on both sides of the measurement equation. This is common in extended object tracking because the measurement source depends on the actual measurement. To implement this equation, we first use the known measurement to determine the measurement source and assume it to be a constant. Then, we perform the update as usual, since the right side does not depend on $\vecz$ anymore.

This equation can be used as the measurement equation in a suitable nonlinear filter. For example, one can use a nonlinear version of the Kalman filter such as the UKF~\cite{julier2004} or the S$^2$KF~\cite{JAIF16_Symmetric_S2KF_Steinbring}.

In the proposed measurement equation, the association problem is solved using a greedy association model (GAM), where each measurement is projected onto the surface of the estimated object along the line from the measurement to the object's star point. This approach works well in most cases. However, it has been shown that greedy association models can cause biased estimates of the object's size if the noise is very large compared to the size of the object~\cite{Fusion14_Faion}. To address this issue, the proposed approach could be extended using the partial information model (PIM) as described in~\cite{Fusion15_Faion} to eliminate the bias.

\section{Simulations}
In this section, we illustrate the proposed method in multiple simulations.

\begin{figure*}
	\centering
%	step 2 iou 0.372414
%	step 3 iou 0.585648
%	step 4 iou 0.675926
%	step 5 iou 0.770833
%	step 21 iou 0.965278
	\begin{tabular}{ccccc}
		IoU = 0.372414 & IoU = 0.585648 &  IoU = 0.675926 &  IoU = 0.770833 &  IoU = 0.965278 \\	
		\includegraphics[width=30mm]{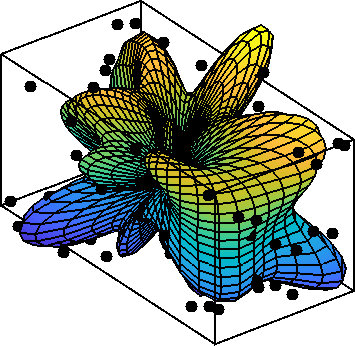} &
		\includegraphics[width=30mm]{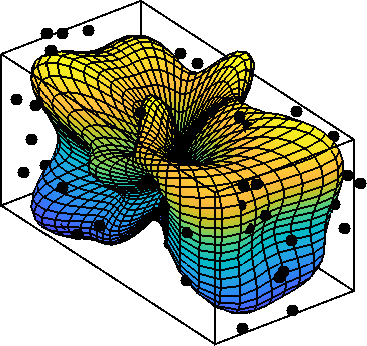} &
		\includegraphics[width=30mm]{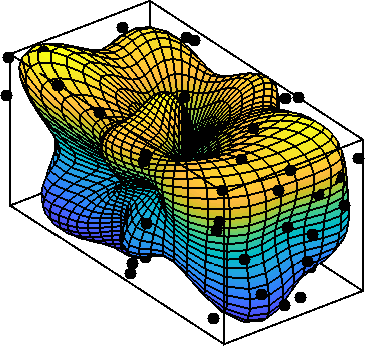} &	
		\includegraphics[width=30mm]{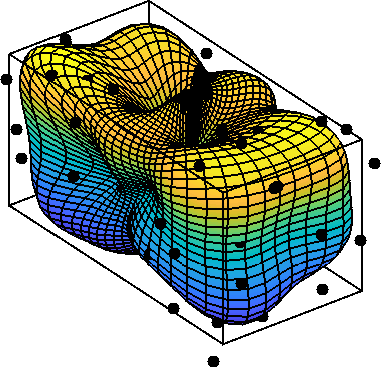} &	
		\includegraphics[width=30mm]{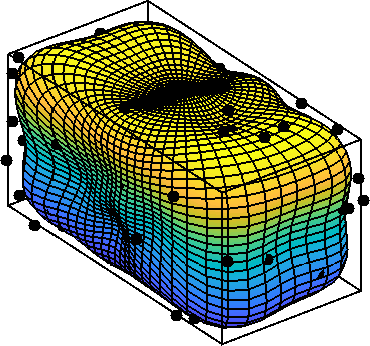} \\
		$k=1$ & $k=2$ & $k=3$ & $k=4$ & $k=20$
	\end{tabular}
	\caption{Estimate of a cuboid with order $L=8$ of spherical harmonics at different time steps.}
	\label{fig:cuboid-convergence}
\end{figure*}

\begin{figure*}
	\centering
%	order 1 iou 0.481928
%	order 2 iou 0.807606
%	order 4 iou 0.865297
%	order 6 iou 0.928241
%	order 8 iou 0.967593
	\begin{tabular}{ccccc}
		IoU = 0.481928 & IoU = 0.807606 &  IoU = 0.865297 &  IoU = 0.928241 &  IoU = 0.967593 \\	
		\includegraphics[width=30mm]{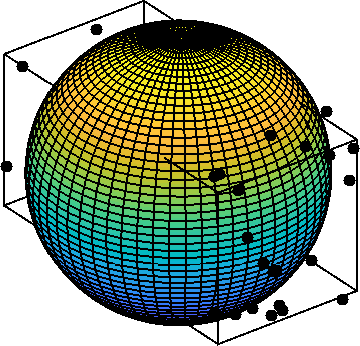} &
		\includegraphics[width=30mm]{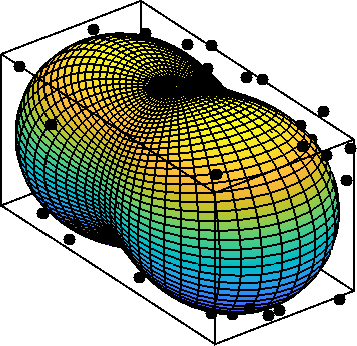} &
		\includegraphics[width=30mm]{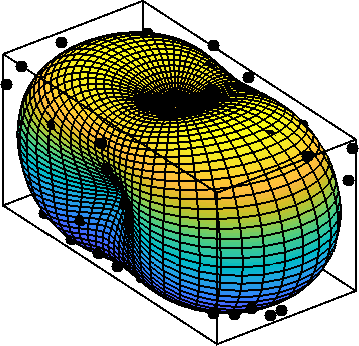} &	
		\includegraphics[width=30mm]{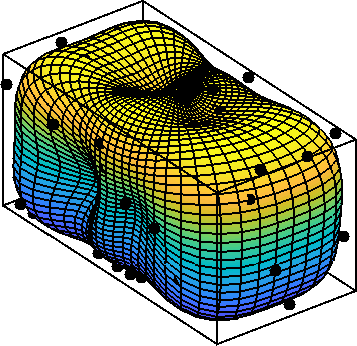} &	
		\includegraphics[width=30mm]{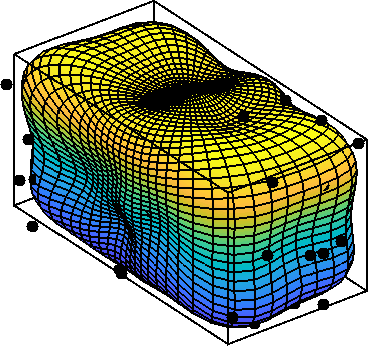} \\
		$L=1$ & $L=2$ & $L=4$ & $L=6$ & $L=8$
	\end{tabular}
	\caption{Estimate of a cuboid with different orders $L$ of spherical harmonics after convergence.}
	\label{fig:cuboid-order}	
\end{figure*}

\subsection{Cuboid}
\newcommand{\mysize}{1}
\begin{figure*}
	\centering
%	step 11 iou 0.878641
%	step 12 iou 0.911111
%	step 13 iou 0.884892
%	step 14 iou 0.894737
%	step 15 iou 0.933002
%	step 11 iou 0.536283
%	step 12 iou 0.541156
%	step 13 iou 0.563884
%	step 14 iou 0.557432
%	step 15 iou 0.546218
	\begin{tabular}{ccccc}
		IoU = 0.878641 & IoU = 0.911111 &  IoU = 0.884892 &  IoU = 0.894737 &  IoU = 0.933002 \\
		\includegraphics[scale=\mysize]{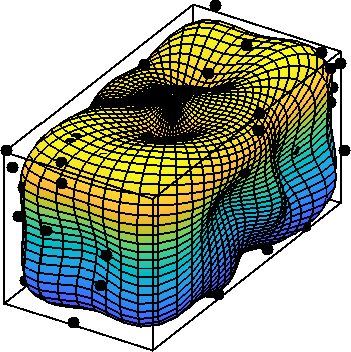} &
		\includegraphics[scale=\mysize]{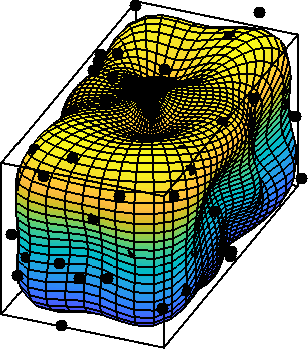} &
		\includegraphics[scale=\mysize]{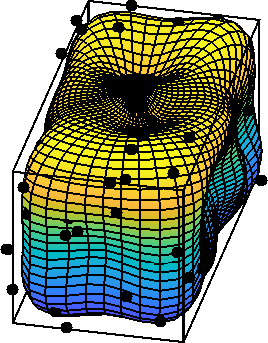} &
		\includegraphics[scale=\mysize]{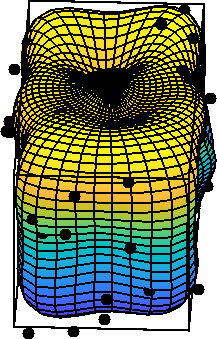} &
		\includegraphics[scale=\mysize]{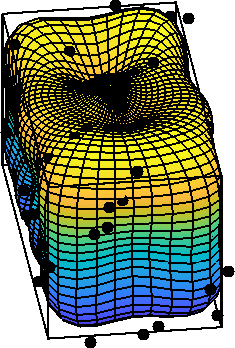} \\
		IoU = 0.536283 & IoU = 0.541156 &  IoU = 0.563884 &  IoU = 0.557432 &  IoU = 0.546218 \\		
		\includegraphics[scale=\mysize]{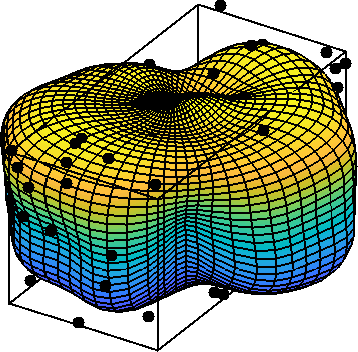} &
		\includegraphics[scale=\mysize]{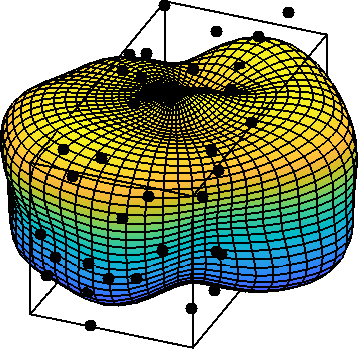} &
		\includegraphics[scale=\mysize]{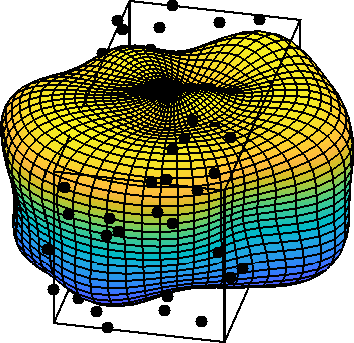} &
		\includegraphics[scale=\mysize]{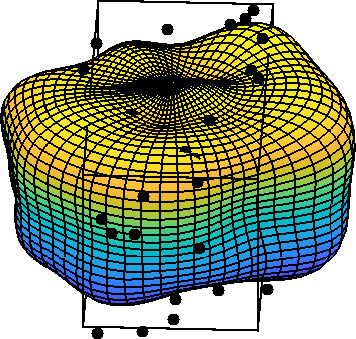} &
		\includegraphics[scale=\mysize]{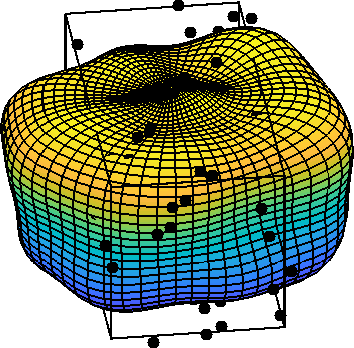} \\	
		$k=11$ & $k=12$ & $k=13$ & $k=14$ & $k=15$
	\end{tabular}
	\caption{Estimate of a rotating cuboid with order $L=8$ at different time steps with motion model (top) and random walk model (bottom).}
	\label{fig:cuboid-rotation}	
\end{figure*}
First, we apply our approach to estimation of a cuboid of size $3 \times 1 \times 1$ units. 
For this purpose, we generate measurements at each time step by sampling a predefined number of random points from the cuboid's surface and adding isotropic Gaussian measurement  noise with a variance of $10^{-2}$. In the following simulations, we always use 60 measurements per time step. Then, we perform recursive estimation using an unscented Kalman filter (UKF). The same filter was used in all simulations and experiments in this paper. To quantify the accuracy of the result, we consider the Intersection over Union (IoU), also known as the Jaccard index~\cite{levandowsky1971}, i.e., the volume of the intersection of the cuboid and the estimated shape divided by the volume of their union. The IoU is always in [0,1], where 1 corresponds to an exact match of the two objects.

In Fig.~\ref{fig:cuboid-convergence}, we depict how the estimated shape converges toward the true shape of the cube over time. The measurements at each time step are shown as black dots. It can be seen that the estimate of the shape is quite poor at first and gets better with every time step. After a few time steps, the shape estimate converges to a fairly accurate representation of the true shape of the cuboid.

It is obvious that the accuracy of the shape estimate depends on the order $L$ of the spherical harmonics, and thus, the number of coefficients used to represent the shape. This effect is illustrated in Fig.~\ref{fig:cuboid-order}, where we show the estimated shape after convergence for different values of $L$. As is obvious, a higher order allows a more accurate representation of the shape, especially around the corners of the cube. 

In order to illustrate the effect of the system model on the estimation results, we perform an experiment with a rotating cuboid. For this experiment, the cube rotates around a fixed axis with an angular rate of 10 degrees per time step. We compare the results from a random walk system model and a system model that considers the angular velocity as a known input and performs an appropriate rotation of the spherical harmonics coefficients according to \cite{blanco1997}. The results of this experiment are shown in Fig.~\ref{fig:cuboid-rotation}. As expected, the shape is estimated much more accurately with the proper system model that considers the rotation. 

\begin{figure*}
	\centering
	\begin{tabular}{ccccc}
		\includegraphics[width=30mm]{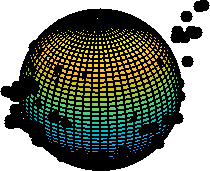} &
		\includegraphics[width=30mm]{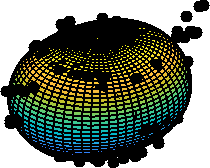} &
		\includegraphics[width=30mm]{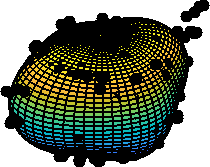} &	
		\includegraphics[width=30mm]{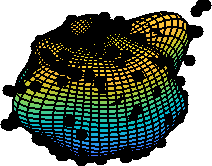} &	
		\includegraphics[width=30mm]{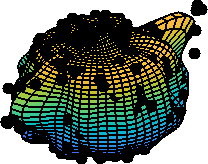} \\
		$L=1$ & $L=2$ & $L=4$ & $L=8$ & $L=12$
	\end{tabular}
	\caption{Estimate of the Utah teapot with different orders $L$ of spherical harmonics after convergence.}
	\label{fig:ply-order}	
\end{figure*}

\begin{figure}
	\centering
	\includegraphics[width=5cm,trim={11cm 1cm 10cm 0cm},clip]{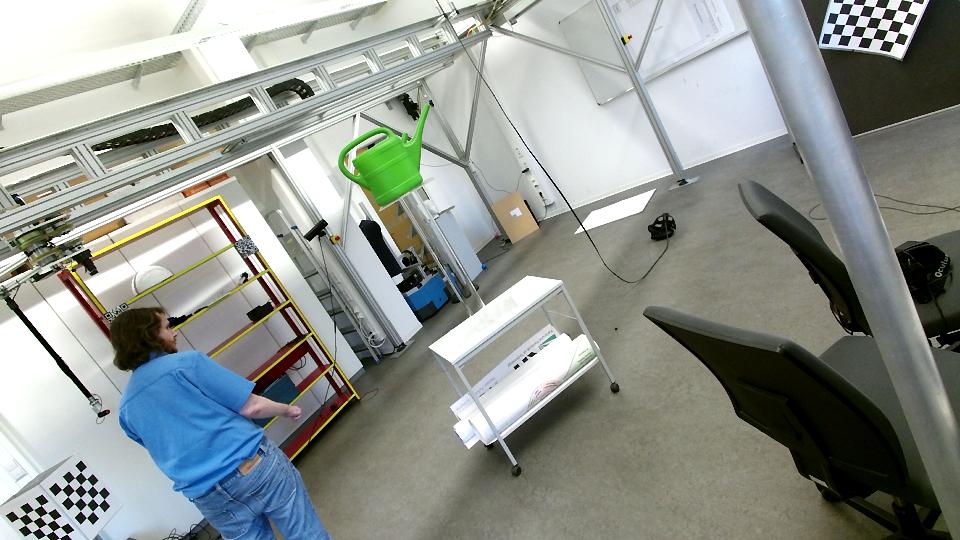}
	\hspace{1cm}
	\includegraphics[width=5cm,trim={11cm 1cm 10cm 0cm},clip]{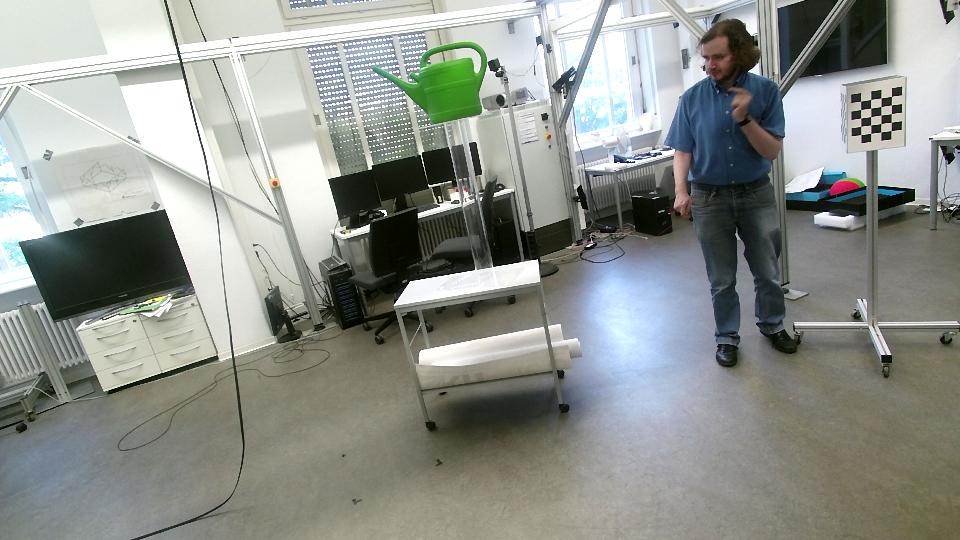}
	\caption{View from both Kinect cameras (cropped to region of interest).}	
	\label{fig:kinect}
\end{figure}

\subsection{Complex 3D Model}
Now, we apply the algorithm used in the previous section to a more complicated 3D model. For this purpose, we use a model of the famous Utah teapot\footnote{Nik Clark, Utah teapot (solid), {https://www.thingiverse.com/thing:852078}}. Our version of the model has a size of about $3.5 \times 2.2 \times 1.8$ units.
For our simulation, we sample 200 random points from the surface of the 3D model at each time step and add isotropic Gaussian measurement noise with a variance of $10^{-2}$ to each point.

The estimated shape after convergence is shown for different numbers of coefficients in Fig.~\ref{fig:ply-order}. Once again, it is clearly visible how a larger number of coefficients (i.e., a higher order) allows representation of more complex shapes.

\section{Experiment}
In order to evaluate the proposed approach in a real scenario, we consider simultaneous tracking and shape estimation of a moving watering can. The watering can is a well-suited object because it has a very characteristic shape and can be approximated quite well using a star-convex model despite possessing non-star convex parts such as the handle.

\begin{figure}
	\centering
	\includegraphics[width=13cm]{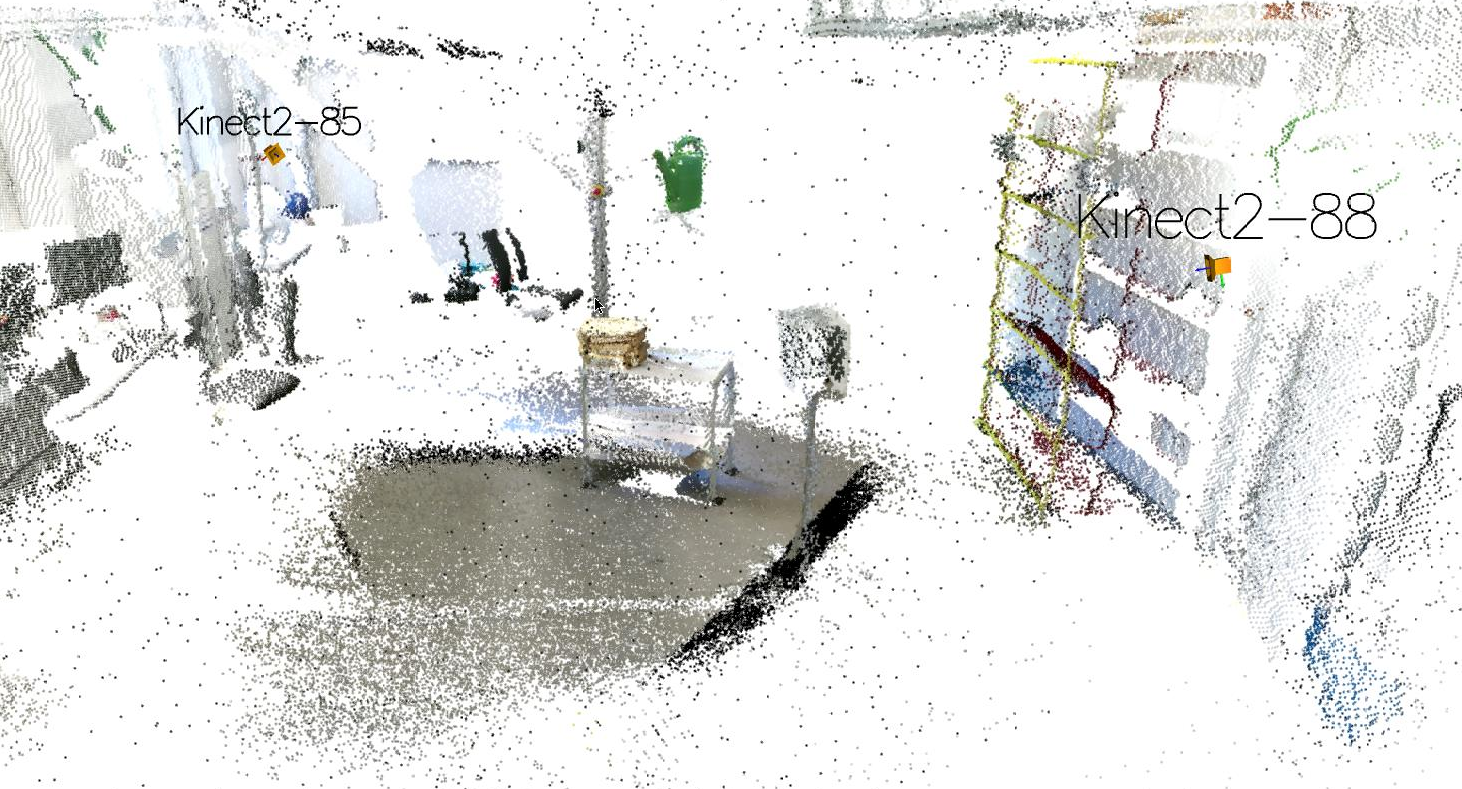}
	\caption{Overview of the experimental setup. One Kinect is located on either side and the watering can to be tracked is placed in the middle.}	
	\label{fig:setup}
\end{figure}

To obtain measurements, we recorded the setting simultaneously with two Kinect 2.0~ \cite{bamji2015} sensors located on opposite sides of the object, such that measurements from both directions can be obtained (see Fig.~\ref{fig:kinect}). These sensors are commonly used in various robotics applications~\cite{yang2015},~\cite{fankhauser2015}. The extrinsic parameters of the Kinect cameras are calibrated using a checkerboard method. Both depth and color images were recorded during the experiment. An overview of our setup is shown in Fig.~\ref{fig:setup}.

We perform segmentation using a bounding box on the 3D points and color thresholding in HSV space. Afterward, we remove outliers that have too few other points close by. This way, we obtain 3D points that belong to the tracked object. Each Kinect records at a frame rate of \SI{30}{\hertz}, so we obtain a total of 60 point clouds per second.
Then, these 3D points are used as measurements for the spherical harmonics algorithm. We assume that the measurements are affected by isotropic Gaussian noise. As an initialization for the location, we use the center of mass of the point cloud at the first time step. The shape is initialized as a sphere, i.e., all $w_l^m$ except $w_0^0$ were zero. For comparison, we implemented the random matrix approach~\cite{feldmann2011}, which is limited to estimation of ellipsoidal shapes.
Furthermore, we implemented the alpha shape approach~\cite{akkiraju1995}, which can be seen as a generalization of the convex hull. However, it does not consider noise and is not able to fuse information recursively.
%https://de.mathworks.com/matlabcentral/fileexchange/28851-alpha-

\begin{figure*}
	\centering
	\newcommand{\figwidth}{50mm}
	\begin{tabular}{rccc}
		\rotatebox{90}{\hspace{10mm}proposed, order 6} &
		\includegraphics[width=\figwidth]{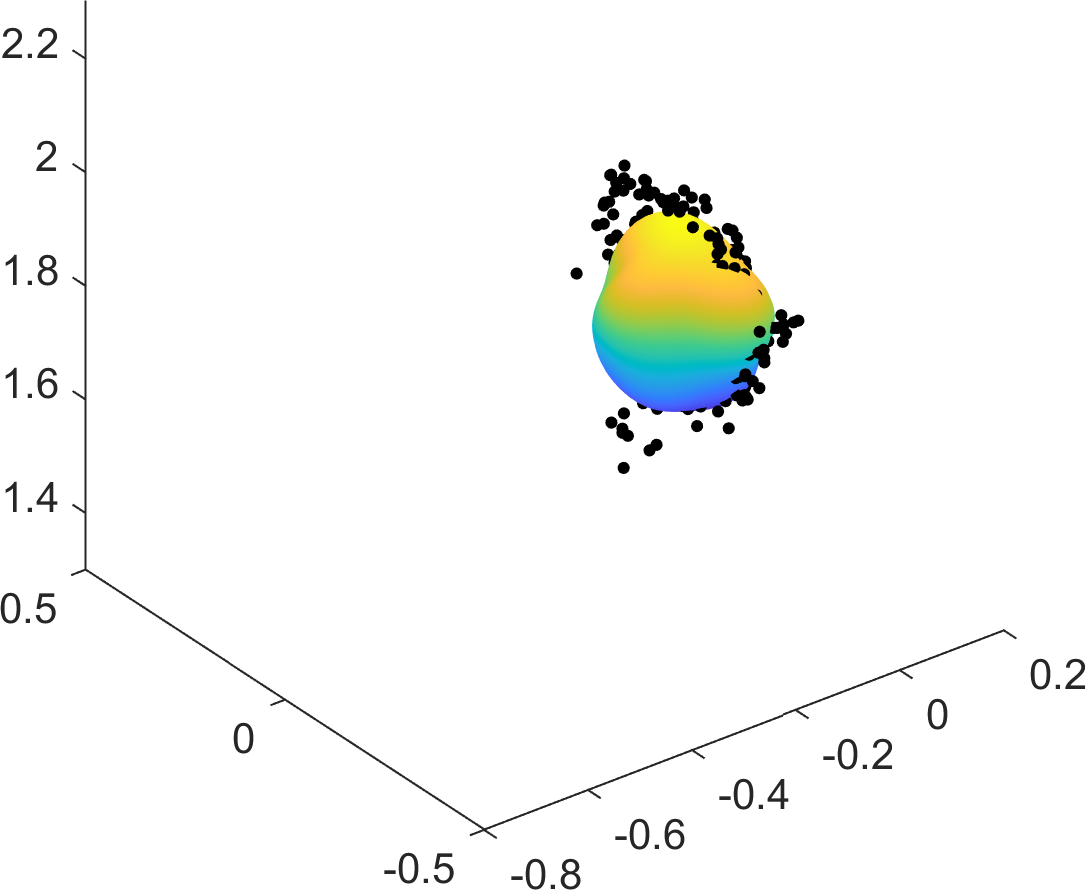} &
		\includegraphics[width=\figwidth]{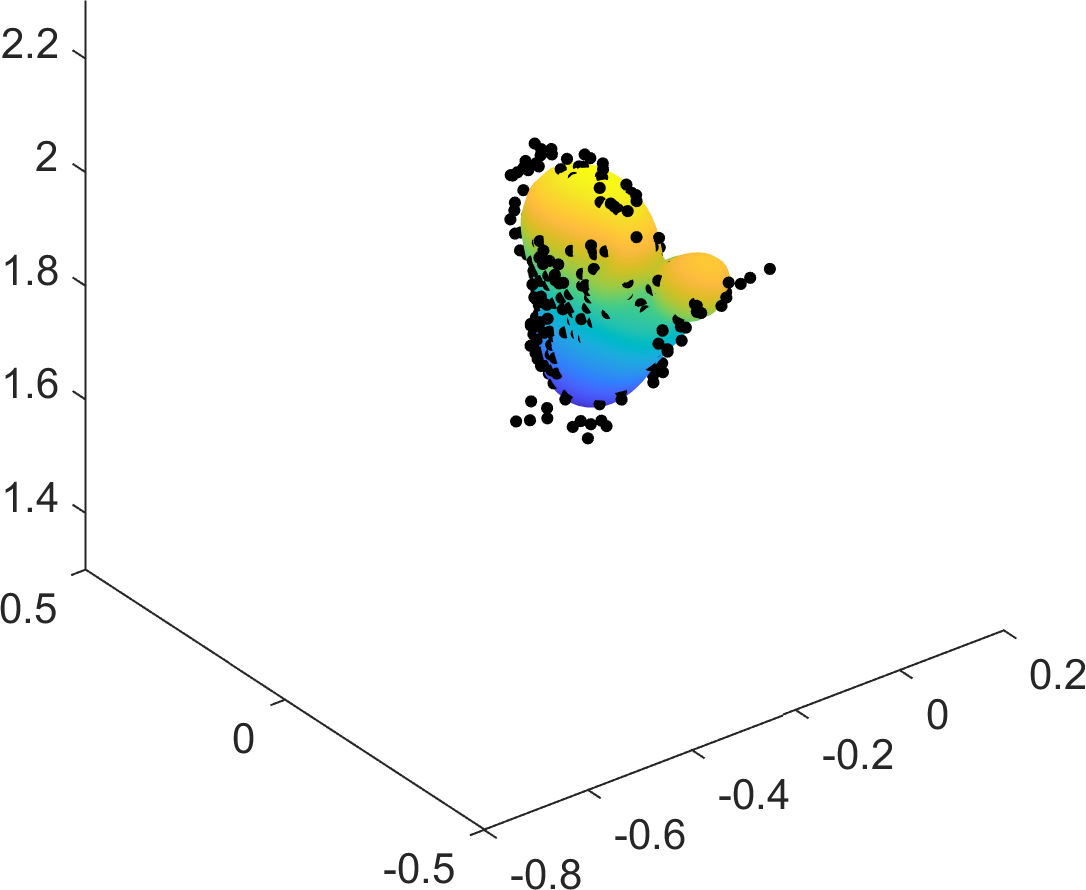} &
		\includegraphics[width=\figwidth]{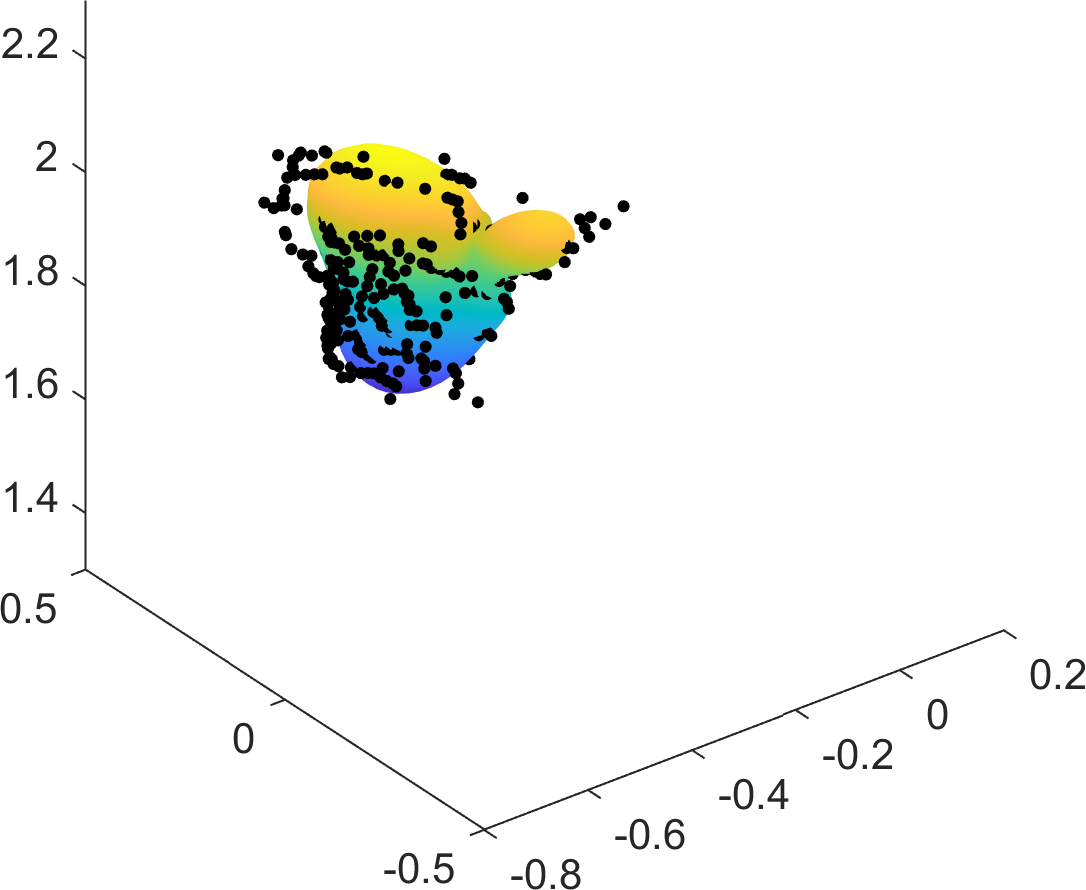} \\	
		\rotatebox{90}{\hspace{10mm}proposed, order 10} &
		\includegraphics[width=\figwidth]{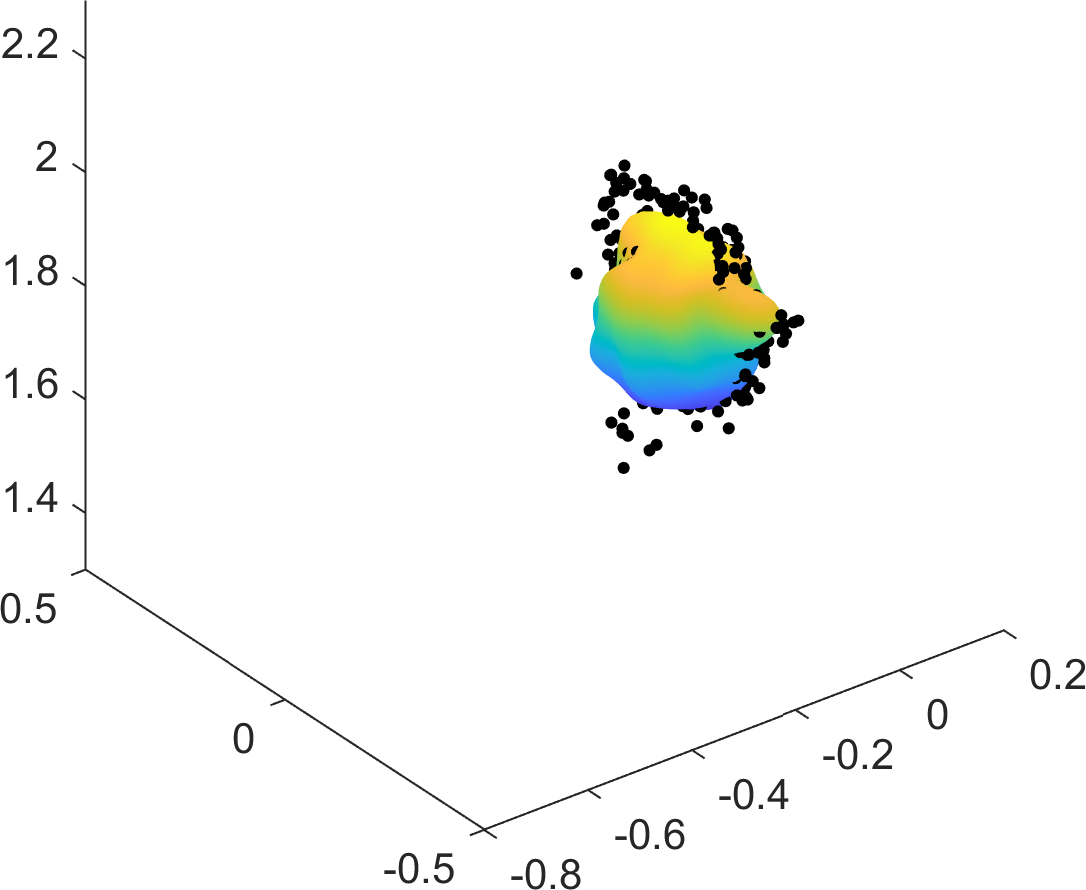} &
		\includegraphics[width=\figwidth]{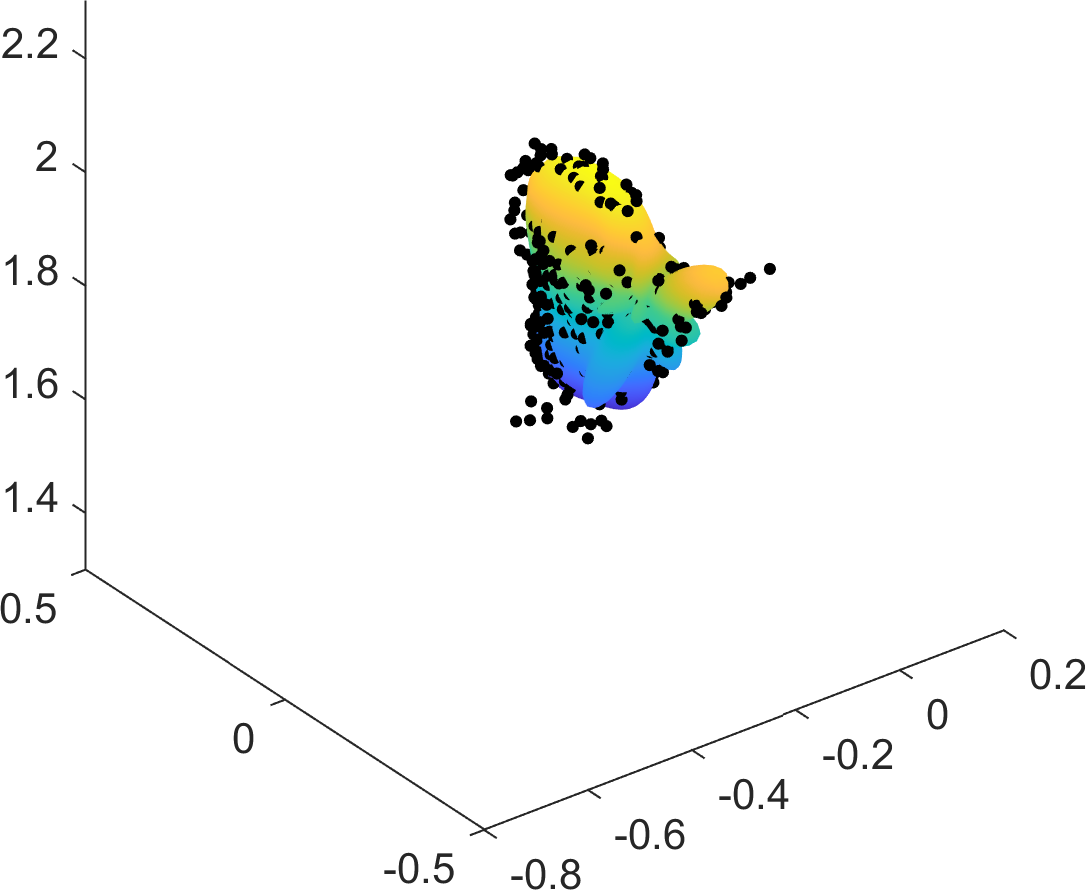} &
		\includegraphics[width=\figwidth]{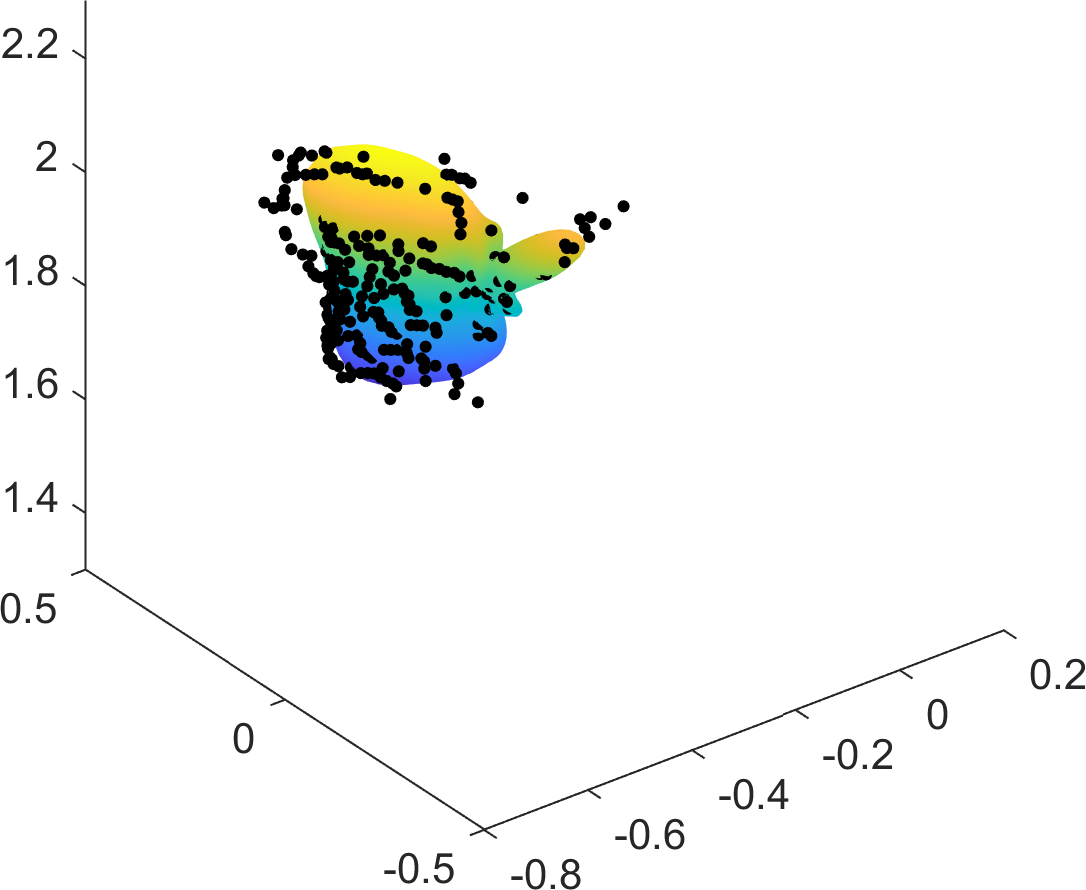} \\	
		\rotatebox{90}{\hspace{10mm}proposed, order 15} &
		\includegraphics[width=\figwidth]{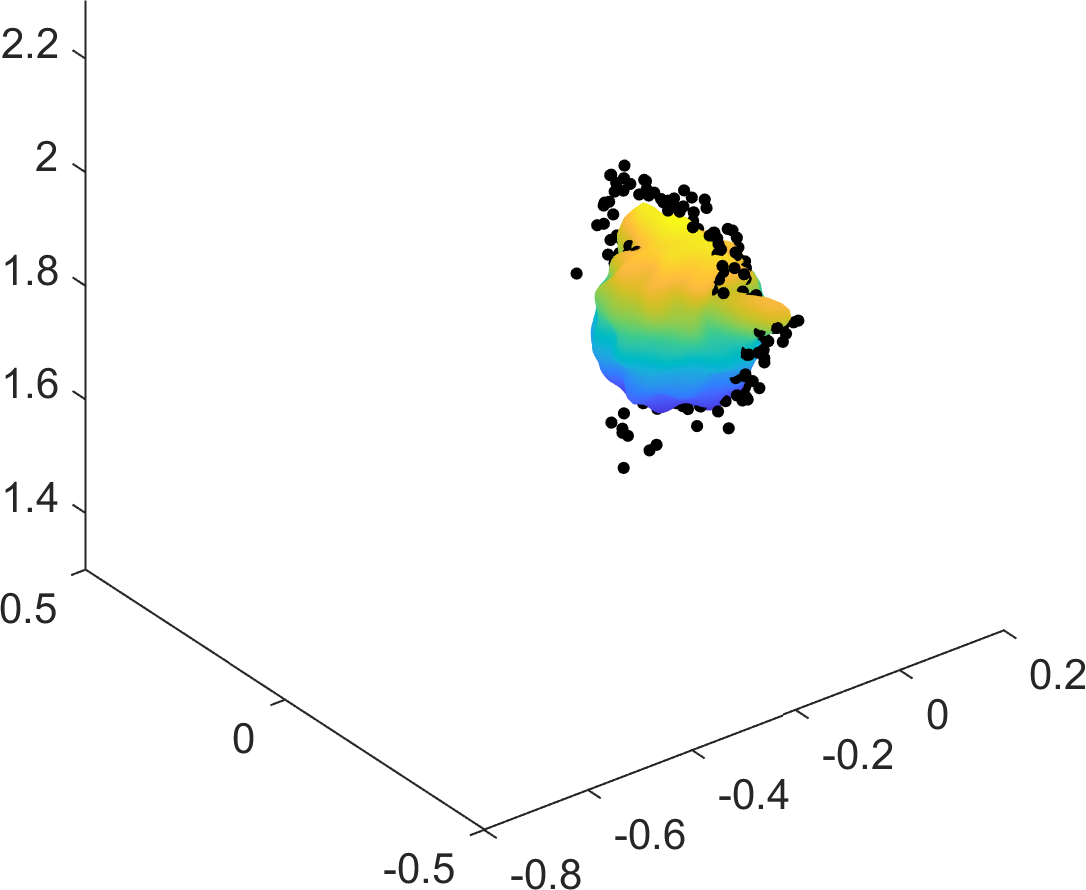} &
		\includegraphics[width=\figwidth]{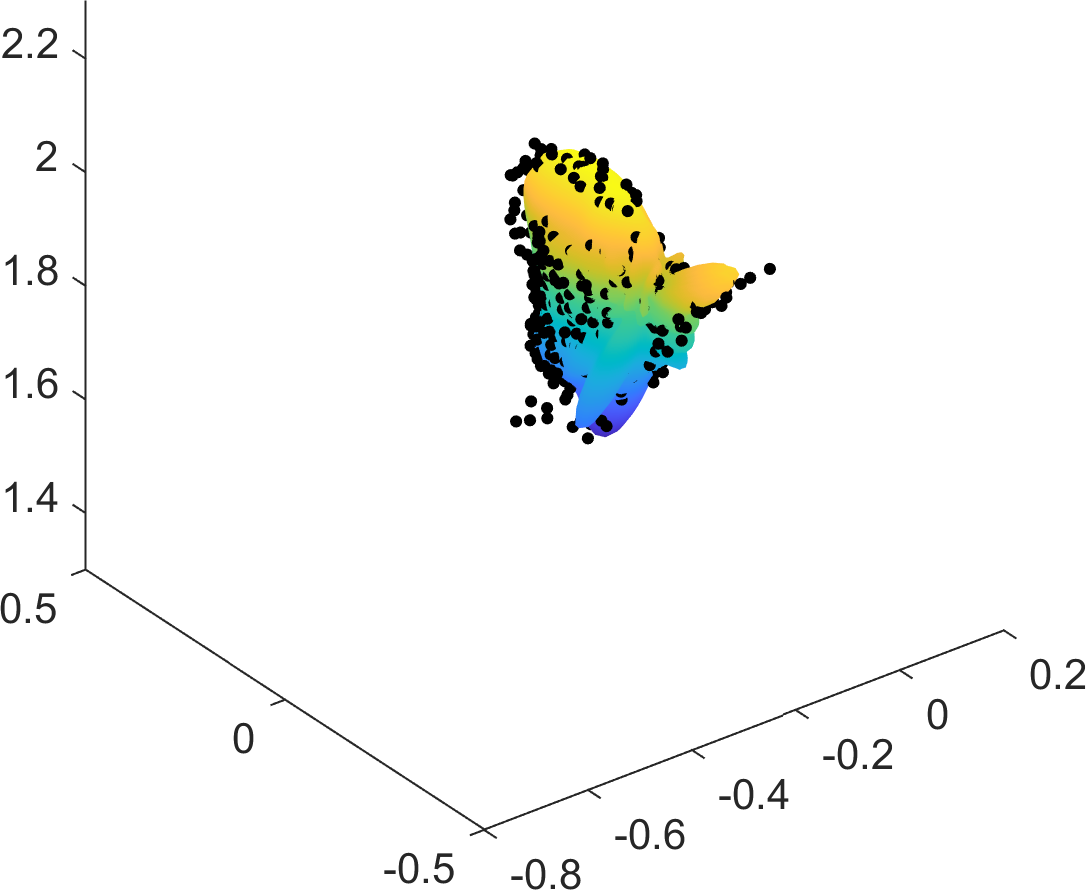} &
		\includegraphics[width=\figwidth]{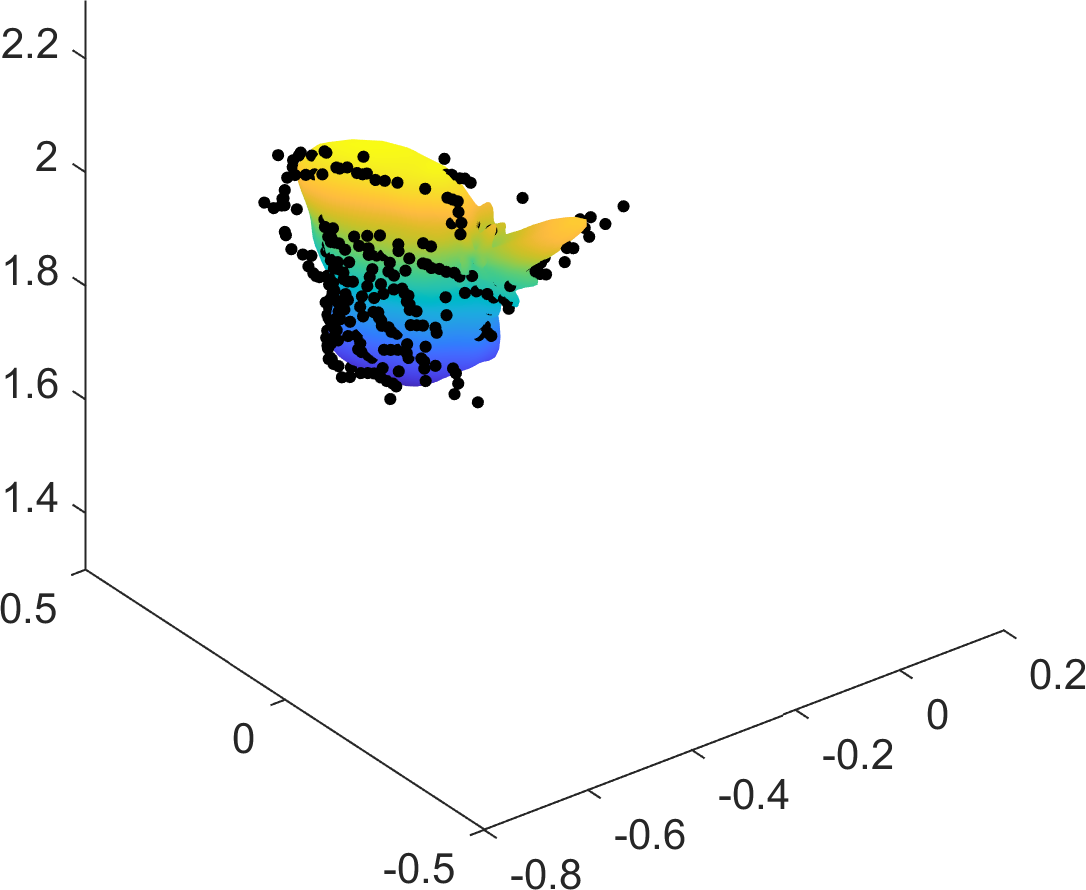} \\
		\rotatebox{90}{\hspace{13mm}random matrices} &
		\includegraphics[width=\figwidth]{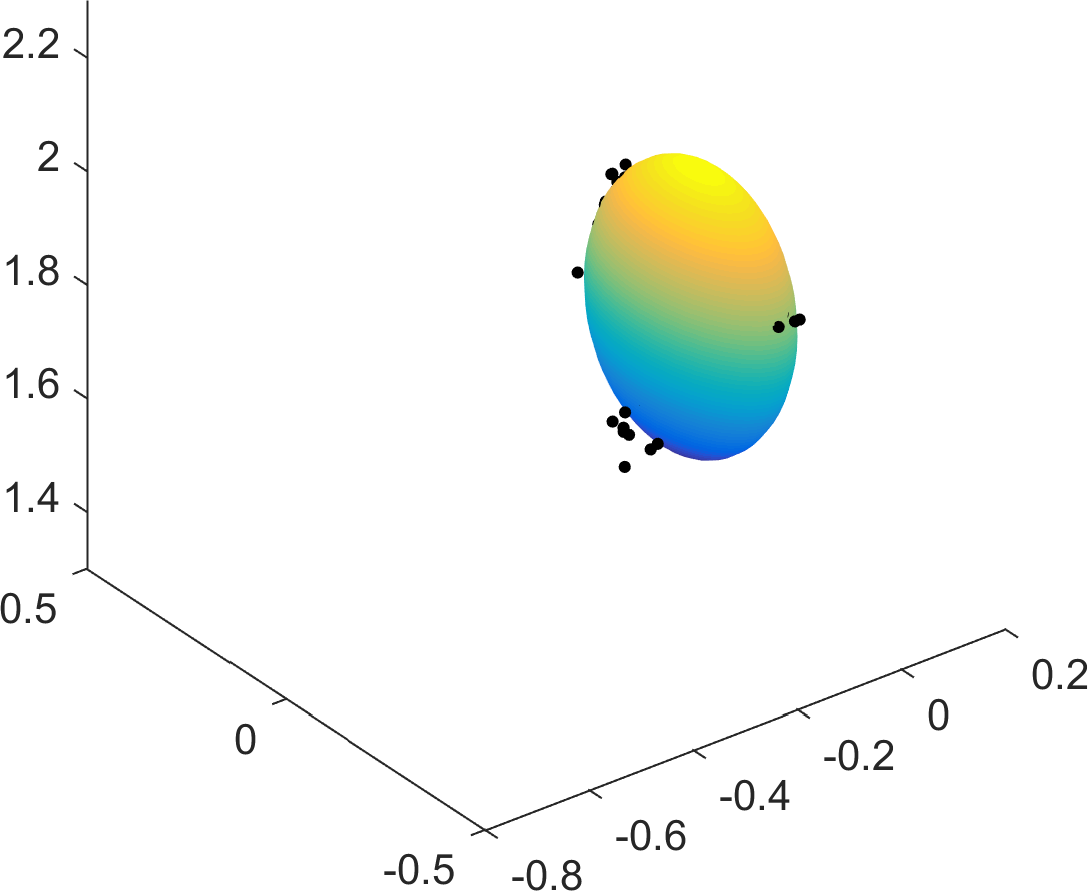} &
		\includegraphics[width=\figwidth]{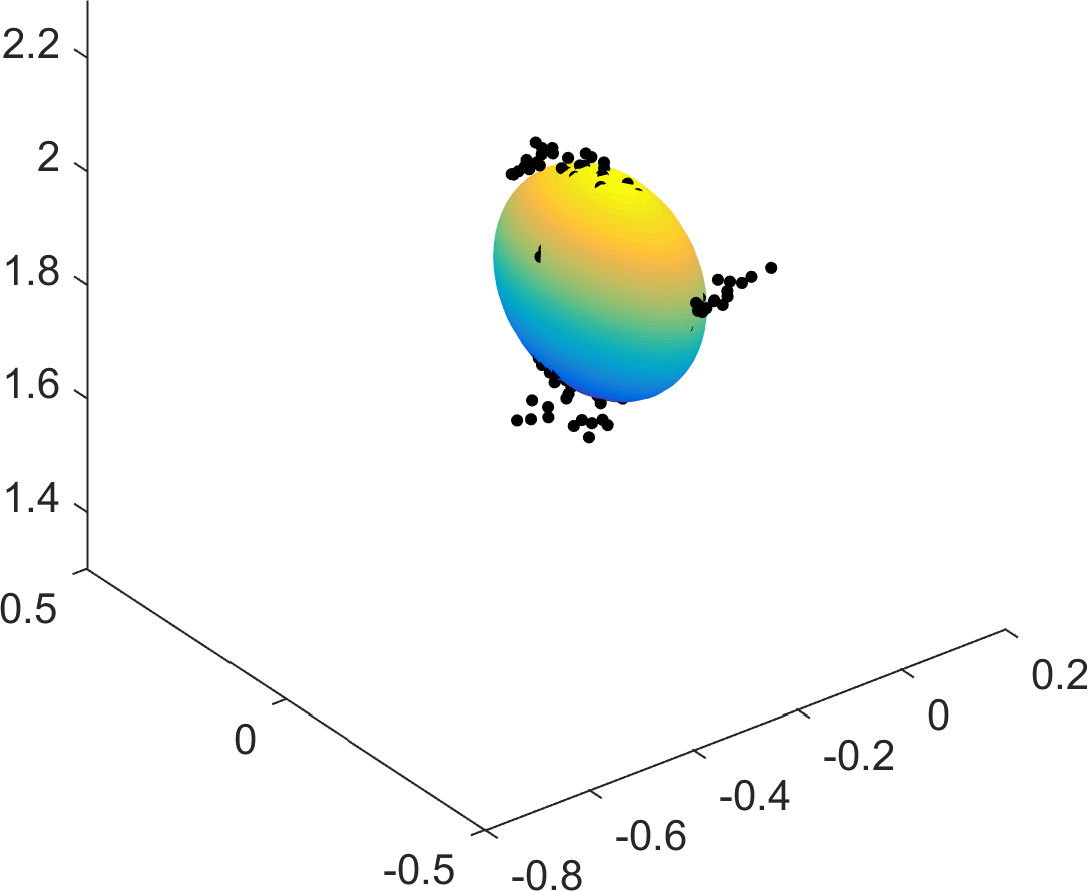} &
		\includegraphics[width=\figwidth]{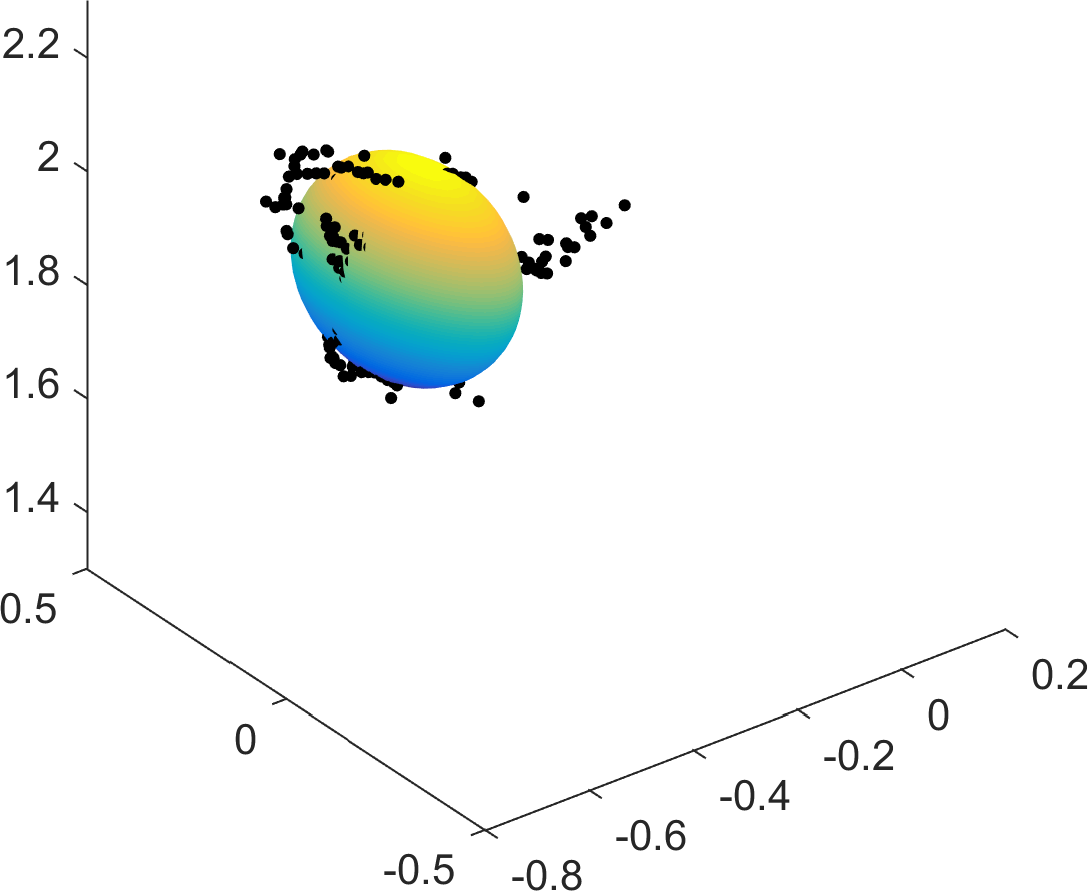} \\		
		\rotatebox{90}{\hspace{16mm}alpha shapes} &
		\includegraphics[width=\figwidth]{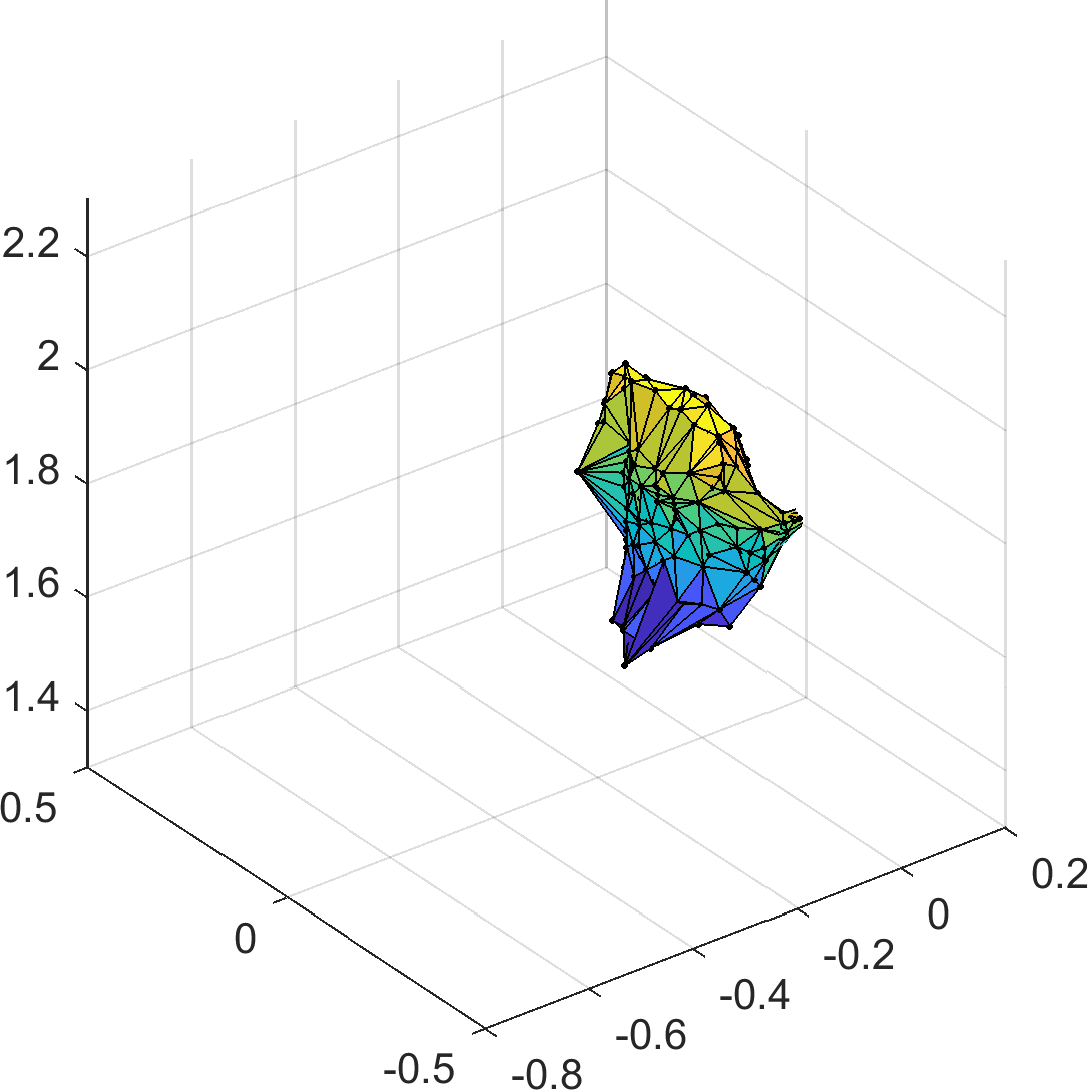} &
		\includegraphics[width=\figwidth]{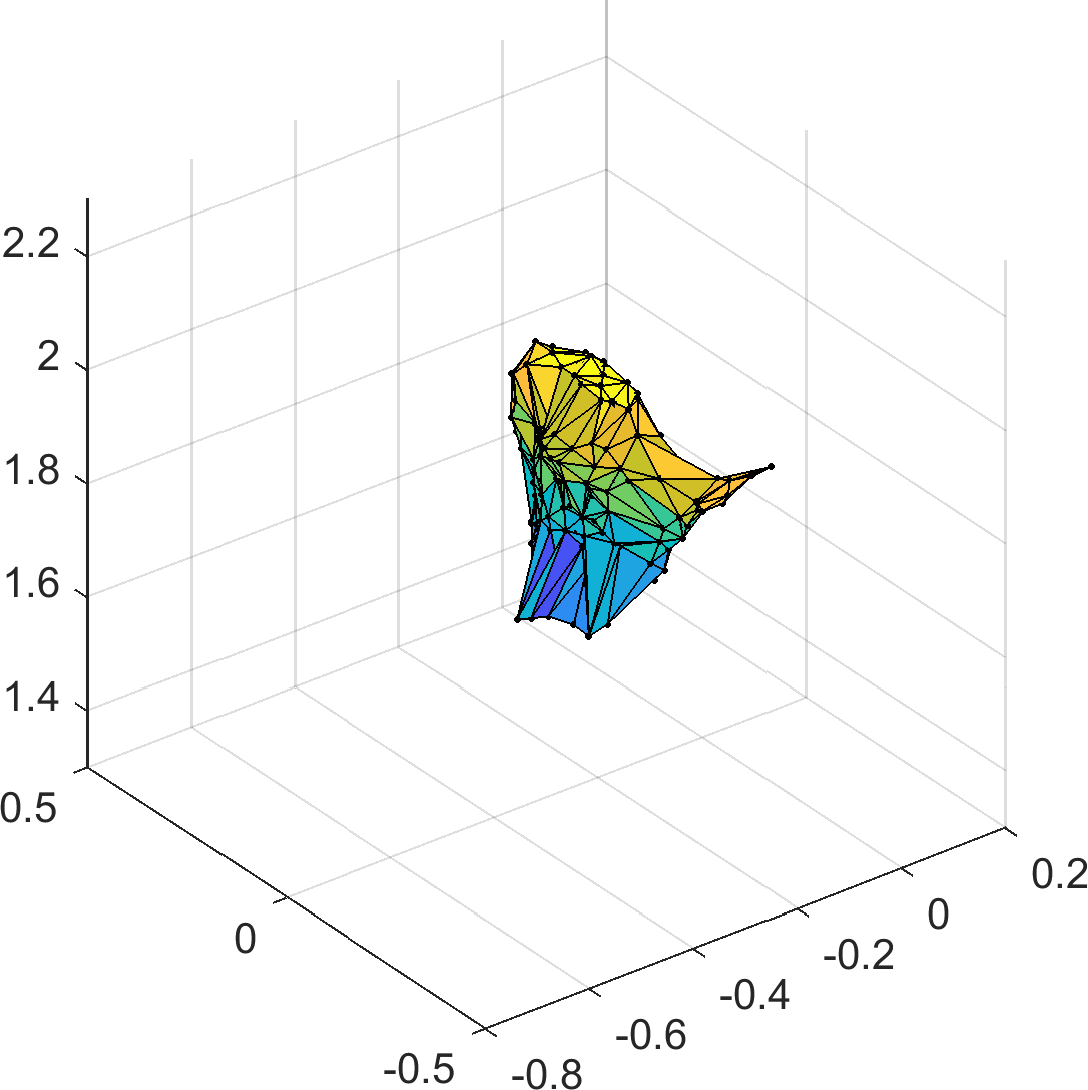} &
		\includegraphics[width=\figwidth]{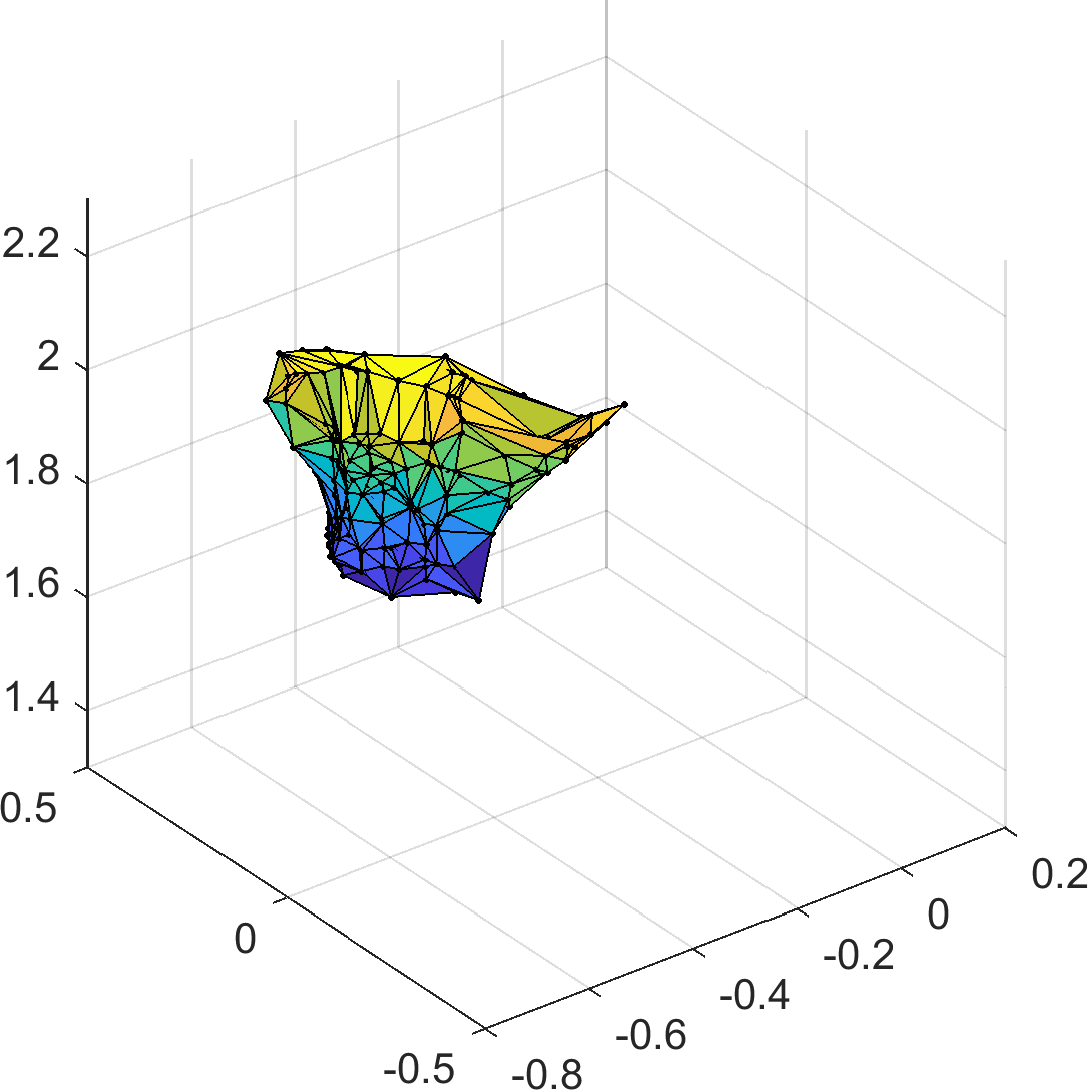} \\			
		& $k=2$ & $k=200$ & $ k=400$
	\end{tabular}
	\caption{Experimental results for order 6 (top row), 10 (second row), and 15 (third row), as well as random matrices (fourth row), and alpha shapes (fifth row) at different time steps. All units are in meters. The measurements of the current time step are shown as black dots.}
	\label{fig:wateringcan}
\end{figure*}

The results of our experiment are shown Fig.~\ref{fig:wateringcan}. It can be seen that a higher number of coefficients once again leads to a more detailed reconstruction. As far as tracking of the location is concerned, all approaches seem to be able to follow the movement of the watering can quite well. The shape estimate of the random matrices approach is very poor as that method is limited to ellipsoidal shapes. While the shape estimate obtained by the alpha volume approach is much more detailed, it heavily suffers from noise and outliers. Also, the estimate can change significantly from one time step to the next.

\section{Conclusion}
We have presented a novel method for extended object tracking with simultaneous shape estimation in three dimensions. The key idea of this method is to represent star-convex objects using spherical harmonics and to estimate their coefficients using standard nonlinear estimation techniques.

Through an evaluation of the proposed method in several simulations as well as a real experiment, we have shown its applicability to a number of different scenarios. The experiments show that a larger number of coefficients allows for a more accurate shape representation. We also illustrate how a known angular velocity can be incorporated. Furthermore, the advantages of the proposed approach compared with random matrices, a method limited to ellipsoidal shapes, as well as alpha shapes has been shown.

Future work may include the use of Gaussian processes~\cite{dragiev2011} instead of spherical harmonics, development of methods that can handle shapes that are not star-convex, and research on methods that can take advantage of symmetries. An extension of the approach based on level sets proposed in~\cite{Fusion13_Zea} to three dimensions might also be interesting to consider. In order to estimate volumetric 3D objects, i.e., objects where measurements are obtained from anywhere within the object rather than just its surface, it would also be possible to combine the random hypersurface model (RHM) idea from~\cite{Fusion11_Baum-RHM} with the spherical harmonics approach.

\section*{Acknowledgment}
This work was partially supported by the German Research Foundation (DFG) under grant HA 3789/13-1.

\bibliographystyle{IEEEtran_Capitalize}
\bibliography{../bib/gk,../bib/ISASPublikationen_laufend,../bib/ISASPublikationen,../bib/ISASPreprints}

\end{document}

%% file: StyleFiles/Abbreviations.tex
\def\vec#1{\underline{#1}}
\def\mat#1{{\mathbf #1}}

\def\1_2{{\frac{1}{2}}}

%% file: StyleFiles/Defs.tex
\def\Box#1{{\fboxsep2pt\fbox{$#1$}}}

\newlength\EqLen

\def\ScaleInner#1{%
\settowidth{\EqLen}{#1}
\ifdim\EqLen < \columnwidth%
  \begin{equation*}%
    \begin{minipage}{\EqLen}#1\end{minipage}%
  \end{equation*}%
\else%
  \begin{equation*}%
    \resizebox{0.99\columnwidth}{!}{\begin{minipage}{\EqLen}#1\end{minipage}}%
  \end{equation*}%
\fi%
}%

\def\Scale#1
  {
  \ScaleInner{$ #1 $} 
  }
  
\def\LongVersion#1{}

\def\citep#1{(\cite{#1})}

\usepackage{amsthm}



\makeatletter
\newcommand\SaveEquation[2]{\@namedef{equation@#1}{#2}}
\newcommand\UseEquation[1]{\@nameuse{equation@#1}}
\makeatother

%% file: StyleFiles/Format.tex
\date{}